\documentclass[10pt,twocolumn,letterpaper]{article}

\usepackage{cvpr}
\usepackage{times}
\usepackage{epsfig}
\usepackage{graphicx}
\usepackage{amsmath}
\usepackage{amssymb}
\usepackage{graphicx}
\usepackage{subfig}
\usepackage{caption}
\usepackage{overpic}
\usepackage{wrapfig}
\usepackage{multirow}

\usepackage{amsthm}
\usepackage{algorithm}
\usepackage{algpseudocode}

\makeatletter
\@namedef{ver@everyshi.sty}{}
\makeatother
\usepackage{tikz}
\usetikzlibrary{arrows, snakes, backgrounds, positioning}
\usepackage{tikzscale}

\newcommand{\R}{\mathbb{R}}

\let\emptyset=\varnothing

\def \saliency {\textup{\saliency}}

\def \path {\textup{path}}
\def \pair {\textup{pair}}
\def \init {\mathit{in}}

\makeatother

\theoremstyle{plain}

\newtheorem{proposition}{\textbf{Proposition}}
\newtheorem{theorem}{\textbf{Theorem}}

\newtheorem{definition}{\textbf{Definition}}
 \theoremstyle{definition}
\numberwithin{theorem}{section}
\numberwithin{lem}{section}

\theoremstyle{remark}\newtheorem{remark}{\textbf{Remark}}

  \let\set=\mathcal 

 \global\long\def\R{\mathbb{R}}

\usepackage[pagebackref=true,breaklinks=true,letterpaper=true,colorlinks,bookmarks=false]{hyperref}

\cvprfinalcopy 

\def\cvprPaperID{3097} 

\ifcvprfinal\fi
\begin{document}

\title{Path-Invariant Map Networks}

\author{Zaiwei Zhang\\
UT Austin
\and
Zhenxiao Liang\\
UT Austin
\and
Lemeng Wu\\
UT Austin
\and
Xiaowei Zhou\\
Zhejiang University\thanks{Xiaowei Zhou is affiliated with the StateKey Lab of CAD\&CG and the ZJU-SenseTime Joint Lab of 3D Vision.}
\and
Qixing Huang\\
UT Austin\thanks{huangqx@cs.utexas.edu}
}

\maketitle

\begin{abstract}
Optimizing a network of maps among a collection of objects/domains (or map synchronization) is a central problem across computer vision and many other relevant fields. Compared to optimizing pairwise maps in isolation, the benefit of map synchronization is that there are natural constraints among a map network that can improve the quality of individual maps. While such self-supervision constraints are well-understood for undirected map networks (e.g., the cycle-consistency constraint), they are under-explored for directed map networks, which naturally arise when maps are given by parametric maps (e.g., a feed-forward neural network). In this paper, we study a natural self-supervision constraint for directed map networks called \textsl{path-invariance}, which enforces that composite maps along different paths between a fixed pair of source and target domains are identical. We introduce \textsl{path-invariance bases} for efficient encoding of the path-invariance constraint and present an algorithm that outputs a path-variance basis with polynomial time and space complexities. We demonstrate the effectiveness of our approach on optimizing object correspondences, estimating dense image maps via neural networks, and semantic segmentation of 3D scenes via map networks of diverse 3D representations. In particular, for 3D semantic segmentation our approach only requires 8\% labeled data from ScanNet to achieve the same performance as training a single 3D segmentation network with 30\% to 100\% labeled data.
\end{abstract}

\section{Introduction}
\label{Section:Introduction}

Optimizing a network of maps among a collection of objects/domains (or map synchronization) is a central problem across computer vision and many other relevant fields. Important applications include establishing consistent feature correspondences for multi-view structure-from-motion~\cite{Agarwal:2011:BRD,DBLP:journals/pami/CrandallOSH13,Snavely:2006:PTE,DBLP:conf/iccv/ChatterjeeG13}, computing consistent relative camera poses for 3D reconstruction~\cite{DBLP:journals/ivc/HuberH03,Huang:2006:RFO}, dense image flows~\cite{DBLP:conf/cvpr/ZhouLYE15,DBLP:conf/cvpr/ZhouKAHE16}, image translation~\cite{DBLP:journals/corr/ZhuPIE17,DBLP:journals/corr/YiZTG17}, and optimizing consistent dense correspondences for co-segmentation~\cite{Wang:2013:ICV,DBLP:journals/tog/HuangWG14,DBLP:conf/cvpr/WangHOG14} and object discovery~\cite{DBLP:conf/cvpr/RubinsteinJKL13,DBLP:conf/cvpr/ChoKSP15}, just to name a few. The benefit of optimizing a map network versus optimizing maps between pairs of objects in isolation comes from the \textsl{cycle-consistency} constraint~\cite{DBLP:journals/cgf/NguyenBWYG11,Huang:2012:OAE,Huang:2013:CSM,Wang:2013:ICV}.
For example, this constraint allows us to replace an incorrect map between a pair of dissimilar objects by composing maps along a path of similar objects~\cite{Huang:2012:OAE}. 
Computationally, state-of-the-art map synchronization techniques~\cite{DBLP:conf/icml/BGHHL18,DBLP:conf/icml/ChenGH14,Huang:2013:CSM,DBLP:conf/nips/HuangLBH17,NIPS2016_6128,DBLP:journals/tog/HuangWG14,Huang:2012:OAE,Kim:2012:ECM,DBLP:conf/cvpr/ZhouLYE15,DBLP:conf/iccv/ZhouZD15,DBLP:conf/icra/LeonardosZD17} employ matrix representations of maps~\cite{Kim:2012:ECM,Huang:2012:OAE,Huang:2013:CSM,DBLP:conf/cvpr/WangHOG14,DBLP:journals/tog/HuangWG14}. This allows us to utilize a low-rank formulation of the cycle-consistency constraint (c.f.~\cite{Huang:2013:CSM}), leading to efficient and robust solutions~\cite{DBLP:journals/tog/HuangWG14,DBLP:conf/iccv/ZhouZD15,NIPS2016_6128,DBLP:conf/nips/HuangLBH17}.  

\begin{figure}
\centering
\begin{overpic}[height=0.6\columnwidth]{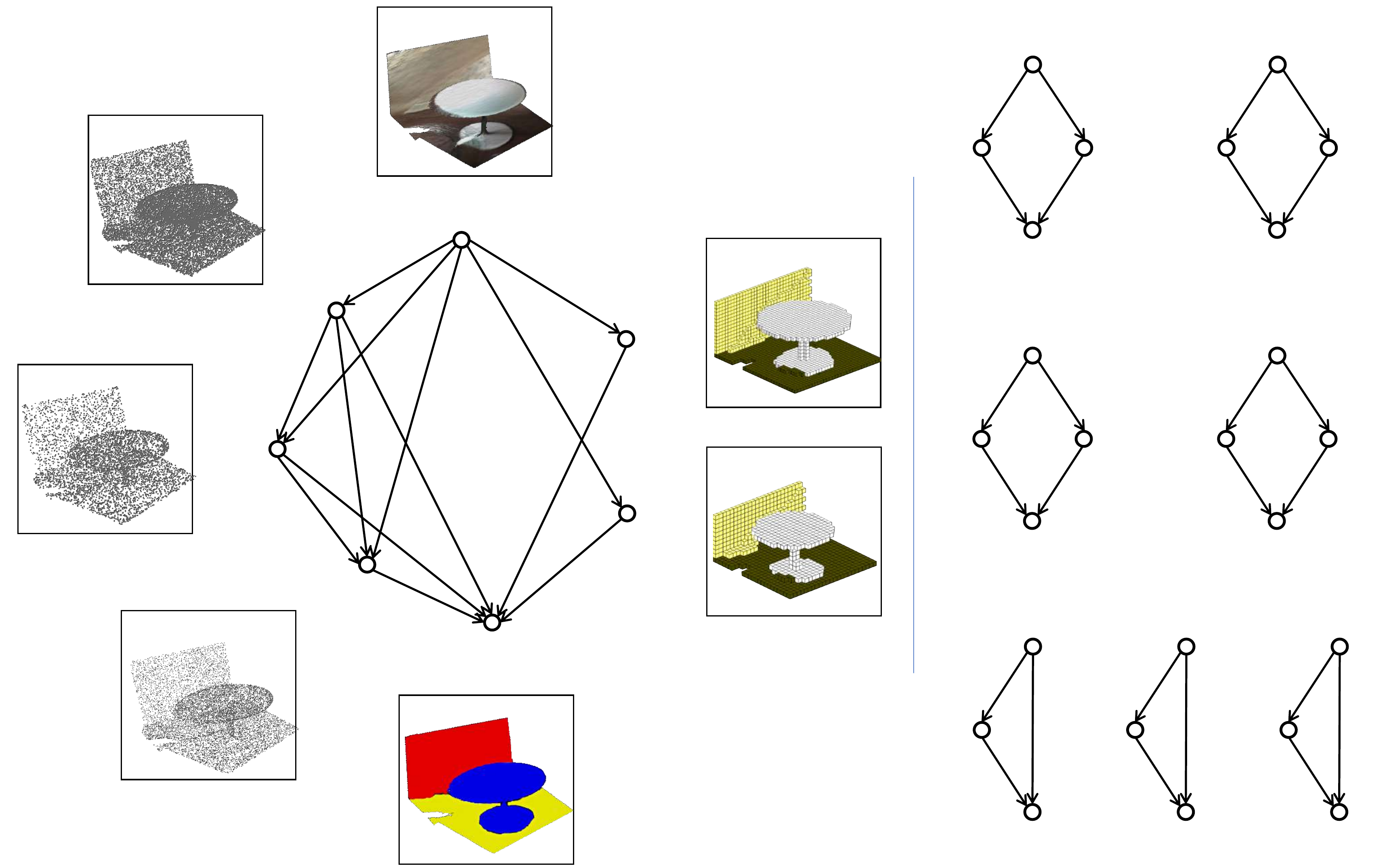}
\put(27,47){\scriptsize{Input Model}}
\put(27,14.5){\scriptsize{Output Seg.}}
\put(20,41){\tiny{PCI}}
\put(15,30){\tiny{PCII}}
\put(21,20){\tiny{PCIII}}
\put(44.2,39.3){\tiny{VOLI}}
\put(44.2,22.2){\tiny{VOLII}}

\put(72,60){\tiny{Input}}
\put(67,52.9){\tiny{PCI}}
\put(78.5,52.9){\tiny{VOLI}}
\put(71.5,43){\tiny{Output}}

\put(90,60){\tiny{Input}}
\put(84.5,52.9){\tiny{PCI}}
\put(96,52.9){\tiny{VOLII}}
\put(89.5,43){\tiny{Output}}

\put(90,39){\tiny{Input}}
\put(84.5,31.9){\tiny{PCI}}
\put(96,31.9){\tiny{PCII}}
\put(89.5,21.5){\tiny{Output}}

\put(72,39){\tiny{Input}}
\put(67,31.9){\tiny{PCI}}
\put(78.5,31.9){\tiny{PCIII}}
\put(71.5,21.5){\tiny{Output}}

\put(71.5,18){\tiny{Input}}
\put(66.5,10.9){\tiny{PCI}}
\put(73,1){\tiny{PCII}}

\put(82.5,18){\tiny{Input}}
\put(77.7,10.9){\tiny{PCI}}
\put(84,1){\tiny{PCIII}}

\put(94.4,17.5){\tiny{PCI}}
\put(88.9,10.9){\tiny{PCII}}
\put(95,1){\tiny{PCIII}}
\end{overpic}
\caption{\small{(Left) A network of 3D representations for the task of semantic segmentation of 3D scenes. (Right) Computed path-invariance basis for regularizing individual neural networks.}}
\label{Figure:Best:Result}
\vspace{-0.2in}
\end{figure}
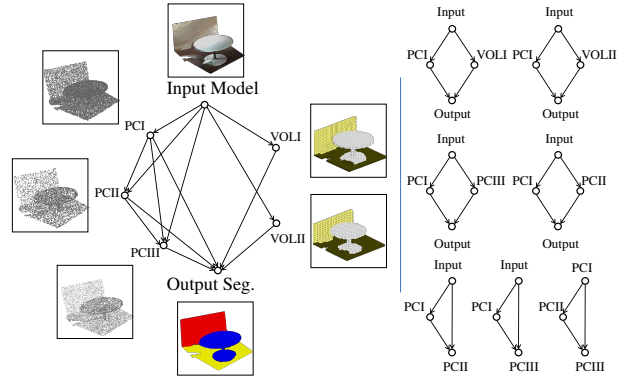

In this paper, we focus on a map synchronization setting, where matrix-based map encodings become too costly or even infeasible. Such instances include optimizing dense flows across many high-resolution images~\cite{Liu:2011:SFD,DBLP:conf/cvpr/KimLSG13,Rubinstein:2016:JIW} or optimizing a network of neural networks, each of which maps one domain to another domain (e.g., 3D semantic segmentation~\cite{dai2017scannet} maps the space of 3D scenes to the space of 3D segmentations). In this setting, maps are usually encoded as broadly defined parametric maps  (e.g., feed-forward neural networks), and map optimization reduces to optimizing hyper-parameters and/or network parameters. Synchronizing parametric maps introduces many technical challenges. For example, unlike correspondences between objects, which are undirected, a parametric map may not have a meaningful inverse map. This raises the challenge of formulating an equivalent regularization constraint of cycle-consistency for directed map networks. In addition, as matrix-based map encodings are infeasible for parametric maps, another key challenge is how to efficiently enforce the regularization constraint for map synchronization. 

We introduce a computational framework for optimizing directed map networks that addresses the challenges described above. Specifically, we propose the so-called \textsl{path-invariance constraint}, which ensures that whenever there exists a map from a source domain to a target domain (through map composition along a path), the map is unique. This path-invariance constraint not only warrants that a map network is well-defined, but more importantly it provides a natural regularization constraint for optimizing directed map networks. To effectively enforce this path-invariance constraint, we introduce the notion of a \textsl{path-invariance basis}, which collects a subset of path pairs that can induce the path-invariance property of the entire map network. We also present an algorithm for computing a path-invariance basis from an arbitrary directed map network. The algorithm possesses polynomial time and space complexities. 

We demonstrate the effectiveness of our approach on three settings of map synchronization. The first setting considers undirected map networks that can be optimized using low-rank formulations~\cite{DBLP:journals/tog/HuangWG14,DBLP:conf/iccv/ZhouZD15}. Experimental results show that our new formulation leads to competitive and sometimes better results than state-of-the-art low-rank formulations. The second setting studies consistent dense image maps, where each pairwise map is given by a neural network. Experimental results show that our approach significantly outperforms state-of-the-art approaches for computing dense image correspondences. The third setting considers a map network that consists of $6$ different 3D representations (e.g., point cloud and volumetric representations) for the task of semantic 3D semantic segmentation (See Figure~\ref{Figure:Best:Result}). By enforcing the path-invariance of neural networks on unlabeled data, our approach only requires 8\% labeled data from ScanNet~\cite{dai2017scannet} to achieve the same performance as training a single semantic segmentation network with 30\% to 100\% labeled data. 

\section{Related Works}
\label{Section:Related:Works}

\noindent\textbf{Map synchronization.} Most map synchronization techniques~\cite{DBLP:journals/ivc/HuberH03,Huang:2006:RFO,conf/cvpr/ZachKP10,DBLP:journals/cgf/NguyenBWYG11,Huang:2013:CSM,DBLP:conf/cvpr/ZhouLYE15,DBLP:journals/tog/HuangWG14,DBLP:conf/icml/ChenGH14,journals/corr/abs-1211-2441,DBLP:conf/iccv/ChatterjeeG13,DBLP:conf/iccv/ZhouZD15,DBLP:journals/corr/ArrigoniFRF15,chen2016PPM,DBLP:conf/nips/HuangLBH17,NIPS2013_4987,NIPS2016_6128,DBLP:conf/icml/HuangCG14,DBLP:conf/cvpr/ZhouKAHE16,DBLP:journals/corr/ZhuPIE17,DBLP:journals/corr/YiZTG17} have focused on undirected map graphs, where the self-regularization constraint is given by cycle-consistency. Depending on how the cycle-consistency constraint is applied, existing approaches fall into three categories. The first category of methods~\cite{DBLP:journals/ivc/HuberH03,Huang:2006:RFO} utilizes the fact that a collection of cycle-consistent maps can be generated from maps associated with a spanning tree. However, it is hard to apply them for optimizing cycle-consistent neural networks, where the neural networks change during the course of the optimization. The second category of approaches \cite{conf/cvpr/ZachKP10,DBLP:journals/cgf/NguyenBWYG11,DBLP:conf/cvpr/ZhouLYE15} applies constrained optimization to select cycle-consistent maps. These approaches are typically formulated so that the objective functions encode the score of selected maps, and the constraints enforce the consistency of selected maps along cycles.
Our approach is relevant to this category of methods but addresses a different problem of optimizing maps along directed map networks. 

The third category of approaches apply modern numerical optimization techniques to optimize cycle-consistent maps. 
Along this line, people have introduced convex optimization~\cite{Huang:2013:CSM,DBLP:journals/tog/HuangWG14,DBLP:conf/icml/ChenGH14,journals/corr/abs-1211-2441}, non-convex optimization~\cite{DBLP:conf/iccv/ChatterjeeG13,DBLP:conf/iccv/ZhouZD15,DBLP:journals/corr/ArrigoniFRF15,chen2016PPM,DBLP:conf/nips/HuangLBH17}, and spectral techniques~\cite{NIPS2013_4987,NIPS2016_6128}. To apply these techniques for parametric maps, we have to hand-craft an additional latent domain, as well as parametric maps between each input domain and this latent domain, which may suffer from the issue of sub-optimal network design. 
In contrast, we focus on directed map networks among diverse domains and explicitly enforce the path-invariance constraint via path-invariance bases. 

\noindent\textbf{Joint learning of neural networks.} Several recent works have studied the problem of enforcing cycle-consistency among a cycle of neural networks for improving the quality of individual networks along the cycle. Zhou et al.~\cite{DBLP:conf/cvpr/ZhouKAHE16} studied how to train dense image correspondences between real image objects through two real-2-synthetic networks and ground-truth correspondences between synthetic images. \cite{DBLP:journals/corr/ZhuPIE17,DBLP:journals/corr/YiZTG17} enforce the bi-directional consistency of transformation networks between two image domains to improve the image translation results. People have applied such techniques for multilingual machine translation~\cite{DBLP:journals/corr/JohnsonSLKWCTVW16}. However, in these works the cycles are explicitly given. In contrast, we study how to extend the cycle-consistency constraint on undirected graphs to the path-invariance constraint on directed graphs. In particular, we focus on how to compute a path-invariance basis for enforcing the path-invariance constraint efficiently. 
A recent work~\cite{DBLP:journals/corr/abs-1804-08328} studies how to build a network of representations for boosting individual tasks. However, self-supervision constraints such as cycle-consistency and path-invariance are not employed. Another distinction is that our approach seeks to leverage unlabeled data, while \cite{DBLP:journals/corr/abs-1804-08328} focuses on transferring labeled data under different representations/tasks. Our approach is also related to model/data distillation (See \cite{data_dis} and references therein), which can be considered as a special graph with many edges between two domains. In this paper, we focus on defining self-supervision for general graphs. 

\noindent\textbf{Cycle-bases.} Path-invariance bases are related to cycle-bases on undirected graphs~\cite{Kavitha:2009:SCB}, in which any cycle of a graph is given by a linear combination of the cycles in a cycle-basis. However, besides fundamental cycle-bases~\cite{Kavitha:2009:SCB} that can generalize to define cycle-consistency bases, it is an open problem whether other types of cycle-bases generalize or not. Moreover, there are fundamental differences between undirected and directed map networks. This calls for new tools for defining and computing path-invariance bases. 

\section{Path-Invariance of Directed Map Networks}

In this section, we focus on the theoretical contribution of this paper, which introduces an algorithm for computing a path-invariance basis that enforces the path-invariance constraint of a directed map network. 
Note that the proofs of theorems and propositions in this section are deferred to the supplementary material. 

\subsection{Path-Invariance Constraint}
\label{Section:Path:Invariance}

We first define the notion of a directed map network:
\begin{definition}
A directed map network $\set{F}$ is an attributed directed graph $\set{G} = (\set{V}, \set{E})$ where $V=\{v_1,\dots,v_{|\set{V}|}\}$. Each vertex $v_i\in \set{V}$ is associated with a domain $\set{D}_i$. Each edge $e\in \set{E}$ with $e=(i,j)$ is associated with a map $f_{ij}:\set{D}_i \rightarrow \set{D}_j$. In the following, we always assume $\set{E}$ contains the self-loop at each vertex, and the map associated with each self-loop is the identity map. 
\label{Def:Map:Network}
\end{definition}
For simplicity, whenever it can be inferred from the context we simplify the terminology of a directed map network as a map network. The following definition considers induced maps along paths of a map network.
\begin{definition}
Consider a path $p = (i_0,\cdots,i_k)$ along $\set{G}$. We define the composite map along $p$ induced from a map network $\set{F}$ on $\set{G}$ as
\begin{equation}
f_{p} = f_{i_{k-1} i_k}\circ \cdots \circ f_{i_0 i_1}.
\label{Def:PathMap}
\end{equation}
We also define $f_{\emptyset}:=I$ where $\emptyset$ can refer to any self-loop.
\end{definition}
In the remaining text, for two successive paths $p$ and $q$, we use $p\sim q$ to denote their composition. 

Now we state the path-invariance constraint.
\begin{definition}
Let $\set{G}_{\path}(u,v)$ collect all paths in $\set{G}$ that connect $u$ to $v$. We define the set of all possible path pairs of $\set{G}$ as
$$
\set{G}_{\pair} = \bigcup_{u,v\in\set{V}}\{(p,q)|p,q\in \set{G}_{\path}(u,v)\}.
$$
We say $\set{F}$ is \textsl{path-invariant} if 
\begin{align}
f_{p} = f_{q}, \qquad \forall (p,q)\in \set{G}_{\pair}.
\label{Eq:Path:Invariance}
\end{align}
\end{definition}
\begin{remark}
It is easy to check that path-invariance induces cycle-consistency (c.f.\cite{Huang:2013:CSM}), but cycle-consistency does not necessarily induce path-invariance. For example, a map network with three vertices $\{a,b,c\}$ and three directed maps $f_{ab}$, $f_{bc}$, and $f_{ac}$ has no-cycle, but one path pair $(f_{bc}\circ f_{ab},f_{ac})$.
\end{remark}


\subsection{Path-Invariance Basis} 
\label{Section:Path:Invariance:Basis}

A challenge of enforcing the path-invariant constraint is that there are many possible paths between each pair of domains in a graph, leading to an intractable number of path pairs. This raises the question of how to compute a path-invariance basis $\set{B}\subset \set{G}_{\pair}$, which is a small set of path pairs that are sufficient for enforcing the path-invariance property of any map network $\set{F}$. To rigorously define path-invariance basis, we introduce three primitive operations on path pairs \textup{merge}, \textup{stitch} and \textup{cut}(See Figure~\ref{Figure:Subsitute:Merge}): 

\begin{figure}[b]
\centering
\includegraphics[width=1\columnwidth]{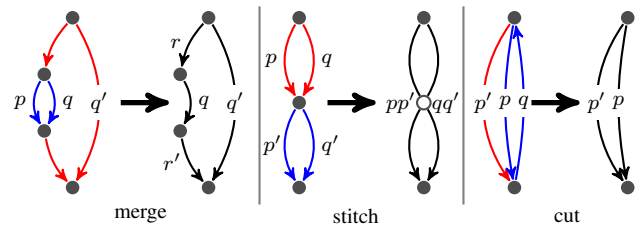}
\caption{Illustrations of Operations}
\label{Figure:Subsitute:Merge}
\label{-0.2in}
\end{figure}
\begin{definition}
Consider a directed graph $\set{G}$. We say two path pairs $(p,q)$ and $(p',q')$ are \textsl{compatible} if one path in $\{p,q\}$ is a sub-path of one path in $\{p',q'\}$ or vice-versa. Without losing generality, suppose $p$ is a sub-path of $p'$ and we write $p' = r\sim p\sim r'$, which stitches three sub-paths $r$,$p$,and $r'$ in order.
We define the \textup{merge} operation so that it takes two compatible path pairs $(p,q)$ and $(r\sim p\sim r',q')$ as input and outputs a new path pair $(r\sim q\sim r',q')$.
\end{definition}
We proceed to define the stitch operation:
\begin{definition} We define the \textup{stitch} operation so that it takes as input two path pairs $(p,q), p,q\in \set{G}_{\path}(u, v)$ and $(p',q'),p',q'\in \set{G}_{\path}(v, w)$ and outputs $(p\sim p',q\sim q')$. 
\end{definition}
Finally we define the cut operation on two cycles, which will be useful for strongly connected graphs:
\begin{definition}
Operation $\textup{cut}$ takes as input two path pairs $(\set{C}_1,\emptyset)$ and $(\set{C}_2,\emptyset)$ where $\set{C}_1$ and $\set{C}_2$ are two distinct cycles that have two common vertices $u,v$ and share a common path from $v$ to $u$. Specifically, we assume these two cycles are $u\xrightarrow{p}v\xrightarrow{q}u$ and $u\xrightarrow{p'}v\xrightarrow{q}u$ where $p,p'\in\set{G}_{\path}(u,v)$ and $q\in\set{G}_{\path}(v,u)$. We define the output of the \textup{cut} operation as a new path pair $(p,p')$. 
\label{def:cut}
\end{definition}
Definition \ref{def:cut} is necessary because $f_{p}\circ f_{q}=f_{p'}\circ f_{q}=I$ implies $f_{p}=f_{p'}$. As we will see later, this operation is useful for deriving new path-invariance basis. 

Now we define path-invariance basis, which is the critical concept of this paper:
\begin{definition}
We say a collection of path pairs $\set{B} = \{(p,q)\}$ is a path-invariance basis on $\set{G}$ if every path-pair $(p,q)\in \set{G}_{\pair}\setminus \set{B}$ can be induced from a subset of $\set{B}$ through a series of \textup{merge}, \textup{stitch} and/or \textup{cut} operations. 
\end{definition}
The following proposition shows the importance of path-invariance basis:
\begin{proposition}
Consider a path-invariance basis $\set{B}$ of a graph $\set{G}$. Then for any map network $\set{F}$ on $\set{G}$, if 
$$
f_{p} = f_{q}, \quad (p,q)\in \set{B},
$$
then $\set{F}$ is path-invariant. 
\end{proposition}

\subsection{Path-Invariance Basis Computation} 
\label{Section:Path:Invariance:Basis:Computation}

We first discuss the criteria for path-invariance basis computation. Since we will formulate a loss term for each path pair in a path-invariance basis, we place the following three objectives. First, we require the length of the paths in each path pair to be small. Intuitively, enforcing the consistency between long paths weakens the regularization on each involved map. Second, we want the size of the resulting path-invariance basis to be small to increase the efficiency of gradient-descent based optimization strategies. Finally, we would like the resulting path-invariance basis to be nicely distributed to improve the convergence property of the induced optimization problem. Unfortunately, achieving these goals exactly appears to be intractable. For example, we conjecture that computing a path-invariance basis of a given graph with minimum size is NP-hard\footnote{Unlike cycle bases that have a known minimum size (c.f.~\cite{Kavitha:2009:SCB}), the sizes of minimum path-invariance bases vary}.

In light of this, our approach seeks to compute a path-invariance basis whose size is polynomial in $|\set{V}|$, i.e., $O(|\set{V}||\set{E}|)$ in the worst case. 
Our approach builds upon the classical result that a directed graph $\set{G}$ can be factored into a directed acyclic graph (or DAG) whose vertices are strongly connected components of $\set{G}$ (c.f.~\cite{books/daglib/0006487}). More precisely, we first show how to compute a path-invariance basis for a DAG. We then discuss the case of strongly connected components. Finally, we show how to extend the result of the first two settings to arbitrary directed graphs. Note that our approach implicitly takes two other criteria into account. Specifically, we argue that small path-invariance basis favors short path-pairs, as it is less likely to combine long path-pairs to produce new path-pairs through merge, stitch and cut operations. In addition, this construction takes the global structure of the input graph $\set{G}$ into account, leading to nicely distributed path-pairs.



\algdef{SE}[SUBALG]{Indent}{EndIndent}{}{\algorithmicend\ }%
\algtext*{Indent}
\algtext*{EndIndent}

\begin{algorithm}[t]
\begin{algorithmic}[1]
\Statex \textbf{input:} Directed graph $\set{G}=(\set{V},\set{E})$.
\Statex \textbf{output:} Path-invariance basis $\set{B}$.
\Indent
    \State Calculate SCCs $\set{G}_1,\dots,\set{G}_K$ for $\set{G}$ and the resulting contracted DAG $\set{G}_{dag}$.
    \State Calculate a path-invriance basis $\set{B}_{dag}$ for $\set{G}_{dag}$ and transform $\set{B}_{dag}$ to $\overline{\set{B}}_{dag}$ that collect path pairs on $\set{G}$.
    \State Calculate a path-invariance basis $\set{B}_{i}$ for $\set{G}_i$.
    \State Calculate path-invirance pairs $\set{B}_{ij}$ whenever $\set{G}_i$ can reach $\set{G}_j$ in $\set{G}_{dag}$.
    \State \Return $\set{B}=\overline{\set{B}}_{dag}\bigcup\big(\cup_{i=1}^K\set{B}_i\big)\bigcup\big(\cup_{ij}\set{B}_{ij}\big)$
\EndIndent
\end{algorithmic}
\caption{The high level algorithm flow to find a path-invariance basis.}
\label{alg:high-level}
\end{algorithm}

\noindent\textbf{Directed acyclic graph (or DAG).} Our algorithm utilizes an important property that every DAG admits a topological order of vertices that are consistent with the edge orientations (c.f.~\cite{books/daglib/0006487}). Specifically, consider a DAG $\set{G} = (\set{V},\set{E})$. A topological order is a bijection $\sigma: \{1,\cdots, |\set{V}|\}\rightarrow \set{V}$ so that we have $\sigma^{-1}(u)<\sigma^{-1}(v)$ whenever $(u,v)\in \set{E}$. A topological order of a DAG can be computed by Tarjan's algorithm (c.f. ~\cite{Tarjan:1976:EST:2696884.2696988}) in linear time. 


Our algorithm starts with a current graph $\set{G}_{cur}=(\set{V},\emptyset)$ to which we add all edges in $\set{E}$ in some order later. Specifically, the edges in $\set{E}$ will be visited with respect to a (partial) edge order $\prec$ where $\forall (u,v),(u',v')\in \set{E}$, $(u,v) \prec (u',v')$ if and only if $\sigma^{-1}(v) < \sigma^{-1}(v')$. Note that two edges $(u,v)$, $(u',v)$ with the same head can be in arbitrary order. 

For each newly visited edge $(u,v)\in \set{E}$, we collect a set of candidate vertices $\set{P}\subset \set{V}$ such that every vertex $w\in \set{P}$ can reach both $u$ and $v$ in $\set{G}_{cur}$. Next we construct a set $\overline{\set{P}}$ by removing from $\set{P}$ all $w\in \set{P}$ such that $w$ can reach some distinct $w'\in \set{P}$. In other words, $w$ is redundant because of $w'$ in this case.
For each vertex $w \in \overline{\set{P}}$, we collect a new path-pair $(p', p\sim  uv)$, where $p$ and $p'$ are shortest paths from $w$ to $u$ and $v$, respectively. After collecting path pairs, we augment $\set{G}_{cur}$ with $(u,v)$. With $\set{B}_{dag}(\sigma)$ we denote the resulting path-pair set after $\set{E}_{cur} = \set{E}$. 



\begin{theorem}
Every topological order $\sigma$ of $\set{G}$ returns a path-invariance basis $\set{B}_{dag}(\sigma)$ whose size is at most $|\set{V}||\set{E}|$.
\end{theorem}


\noindent\textbf{Strongly connected graph (or SCG).} To construct a path-invariance basis of a SCG $\set{G}$, we run a slightly-modified depth-first search on $\set{G}$ from arbitrary vertex. Since $\set{G}$ is strongly connected, the resulting spanning forest must be a tree, denoted by $\set{T}$. The path pair set $\set{B}$ is the result we obtain. In addition, we use a $\set{G}_{dag}$ to collect a acyclic sub-graph of $\set{G}$ and initially it is set as empty. When traversing edge $(u,v)$, if $v$ is visited for the first time, then we add $(u,v)$ to both $\set{T}$ and $\set{G}_{dag}$. Otherwise, there can be two possible cases:

\begin{itemize}
    \item $v$ is an ancestor of $u$ in $\set{T}$. In this case we add cycle pair $(P\sim (u,v),\emptyset)$,  where $P$ is the tree path from $v$ to $u$, into $\set{B}$.
    \item Otherwise, add $(u,v)$ into $\set{G}_{dag}$.
\end{itemize}

We can show that $\set{G}_{dag}$ is indeed an acyclic graph (See Section~\ref{Subsection:Proof:Proposition:2} in the supplementary material). Thus we can obtain a path-invariance basis on $\set{G}_{dag}$ by running the construction procedure introduced for DAG. We add this basis into $\set{B}$. 

\begin{proposition}
The path pair set $\set{B}$ constructed above is a path-invariance basis of $\set{G}$.
\end{proposition}

\noindent\textbf{General directed graph.} Given path-invariance bases constructed on DAGs and SCGs, constructing path-invariance bases on general graphs is straight-forward. Specifically, consider strongly connected components $\set{G}_i, 1\leq i \leq K$ of a graph $\set{G}$. With $\set{G}_{dag}$ we denote the DAG among $\set{G}_i, 1\leq i \leq K$. We first construct path-invariance bases $\set{B}_{dag}$ and $\set{B}_i$ for $\set{G}_{dag}$ and each $\set{G}_i$, respectively. We then construct a path-invariance basis $\set{B}$ of $\set{G}$ by collecting three groups of path pairs. The first group simply combines $\set{B}_i, 1\leq i \leq K$. The second group extends $\set{B}_{dag}$ to the original graph. This is done by replacing each edge $(\set{G}_i, \set{G}_j)\in \set{E}_{dag}$ through a shortest path on $\set{G}$ that connects the representatives of $\set{G}_i$ and $\set{G}_j$ where representatives are arbitrarily chosen at first for each component. To calculate the third group, consider all oriented edges between each  $(\set{G}_i, \set{G}_j)\in \set{E}_{dag}$:
$$
\set{E}_{ij} = \{uv\in\set{E}:u \in  \set{V}_i, v \in \set{V}_j\}.
$$
Note that when constructing $\set{B}_{dag}$, all edges in $\set{E}_{ij}$ are shrinked to one edge in $\set{E}_{dag}$. This means when constructing $\set{B}$, we have to enforce the consistency among $\set{E}_{ij}$ on the original graph $\set{G}$. This can be done by constructing a tree $\set{T}_{ij}$ where $\set{V}(\set{T}_{ij})=\set{E}_{ij}$, $\set{E}(\set{T}_{ij})\subset \set{E}_{ij}^2$. $\set{T}_{ij}$ is a minimum spanning tree on the graph whose vertex set is $\set{E}_{ij}$ and the weight associated with edge $(uv,u'v')\in\set{E}_{ij}^2$ is given by the sum of lengths of $\widehat{uu'}$ and $\widehat{vv'}$. This strategy encourages reducing the total length of the resulting path pairs in $\set{B}_{ij}$:
$$
\set{B}_{ij} := \{(\widehat{uu'}\sim u'v',uv\sim \widehat{vv'}):(uv,u'v')\in \set{E}(\set{T}_{ij})\},
$$
where $\widehat{u u'}$ and $\widehat{v v'}$ denote the shortest paths from $u$ to $u'$ on $\set{G}_i$ and from $v$ to $v'$ on $\set{G}_j$, respectively.
Algorithm~\ref{alg:high-level} presents the high-level pesudo code of our approach. 



\begin{theorem}
The path-pairs $\set{B}$ derived from $\set{B}_{dag}$, \mbox{$\{\set{B}_i:1\leq i\leq K\}$}, and $\{\set{B}_{ij}:(\set{G}_i, \set{G}_j)\in \set{E}_{dag}\}$ using the algorithm described above is a path-invariance basis for $\set{G}$.
\end{theorem}

\begin{proposition}
The size of $\set{B}$ is upper bounded by $|\set{V}||\set{E}|$. 
\end{proposition}

\section{Joint Map Network Optimization}
\label{Section:Approach}

In this section, we present a formulation for jointly optimizing a map network using the path-variance basis computed in the preceding section.

Consider the map network defined in Def.~\ref{Def:Map:Network}. We assume the map associated with each edge $(i,j)\in \set{E}$ is a parametric map $f_{ij}^{\theta_{ij}}$, where $\theta_{ij}$ denotes hyper-parameters or network parameters of $f_{ij}$. We assume the supervision of map network is given by a superset $\overline{\set{E}}\supset \set{E}$. As we will see later, such instances happen when there exist paired data between two domains, but we do not have a direct neural network between them. To utilize such supervision, we define the induced map along an edge $(i,j)\in \overline{\set{E}}$ as the composition map (defined in (\ref{Def:PathMap})) $f_{\widehat{v_i v_j}}^{\Theta}$ along the short path $\widehat{v_i v_j}$ from $v_i$ to $v_j$. Here $\Theta = \{\theta_{ij},(i,j)\in \set{E}\}$ collects all the parameters. We define each supervised loss term as $l_{ij}(f_{ij}^{\Theta}),\forall (i,j)\in \overline{\set{E}}$. The specific definition of $l_{ij}$ will be deferred to Section~\ref{Section:Experimental:Results}.

Besides the supervised loss terms, the key component of joint map network optimization utilizes a self-supervision loss induced from the path-invariance basis $\set{B}$. Let $d_{\set{D}_i}(\cdot,\cdot)$ be a distance measure associated with domain $\set{D}_i$. Consider an empirical distribution $P_i$ of $\set{D}_i$. We define the total loss objective for joint map network optimization as
\begin{equation}
\min\limits_{\Theta} \ \sum\limits_{(i,j)\in \overline{\set{E}}}l_{ij}(f_{\widehat{v_i v_j}}^{\Theta}) + \lambda \sum\limits_{(p,q)\in \set{B}}\underset{v\sim P_{p_t}}{E} d_{\set{D}_{p_t}}(f_{p}^{\Theta}(v), f_{q}^{\Theta}(v))
\label{Eq:Total:Obj}
\end{equation}
where $p_t$ denotes the index of the end vertex of $p$. Essentially, (\ref{Eq:Total:Obj}) combines the supervised loss terms and an unsupervised regularization term that ensures the learned representations are consistent when passing unlabeled instances across the map network. We employ the ADAM optimizer~\cite{DBLP:journals/corr/KingmaB14} for optimization. In addition, we start with a small value of $\lambda$, e.g., $\lambda = 10^{-2}$, to solve (\ref{Eq:Total:Obj}) for 40 epochs. We then double the value of $\lambda$ every 10 epochs. We stop the training procedure when $\lambda \geq 10^3$. The training details are deferred to the Appendix.

\section{Experimental Evaluation}
\label{Section:Experimental:Results}

This section presents an experimental evaluation of our approach across three settings, namely, shape matching (Section~\ref{Section:Network:Object:Maps}), dense image maps (Section~\ref{Section:Network:Dense:Image:Flows}), and 3D semantic segmentation (Section~\ref{Section:Network:3D:Representation}). 

\subsection{Map Network of Shape Maps}
\label{Section:Network:Object:Maps}

We begin with the task of joint shape matching~\cite{DBLP:journals/cgf/NguyenBWYG11,Kim:2012:ECM,Huang:2013:CSM,DBLP:journals/tog/HuangWG14,DBLP:journals/cgf/CosmoRAMC17}, which seeks to jointly optimize shape maps to improve the initial maps computed between pairs of shapes in isolation. We utilize the functional map representation described in~\cite{Ovsjanikov:2012:FMF,Wang:2013:ICV,DBLP:journals/tog/HuangWG14}. Specifically, each domain $\set{D}_i$ is given by a linear space spanned by the leading $m$ eigenvectors of a graph Laplacian~\cite{DBLP:journals/tog/HuangWG14} (we choose $m=30$ in our experiments). The map from $\set{D}_i$ to $\set{D}_j$ is given by a matrix $X_{ij}\in \R^{m\times m}$. Let $\set{B}$ be a path-invariance basis for the associated graph $\set{G}$. Adapting (\ref{Eq:Total:Obj}), we solve the following optimization problem for joint shape matching:
\begin{equation}
\sum\limits_{(i,j)\in \set{E}}\|X_{ij}-X_{ij}^{\init}\|_1 + \lambda \sum\limits_{(p,q)\in \set{B}}\|X_{p} - X_{q}\|_{\set{F}}^2    
\label{Eq:FMap:Opt}
\end{equation}
where 
$\|\cdot\|_1$ and $\|\cdot\|_{\set{F}}$ are the element-wise L1-norm and the matrix Frobenius norm, respectively. $X_{ij}^{\init}$ denotes the initial functional map converted from the corresponding initial shape map using \cite{Ovsjanikov:2012:FMF}.

\noindent\textbf{Dataset.} We perform experimental evaluation on SHREC07--Watertight~\cite{Giorgi_shaperetrieval}. Specifically, SHREC07-Watertight contains 400 shapes across 20 categories. Among them, we choose 11 categories (i.e., Human, Glasses,  Airplane, Ant, Teddy, Hand, Plier, Fish, Bird, Armadillo, Fourleg) that are suitable for inter-shape mapping. We also test our approach on two large-scale datasets Aliens (200 shapes) and Vase (300 shapes) from ShapeCOSEG~\cite{Wang:2012:ACS}. For initial maps, we employ blended intrinsic maps~\cite{Kim:2011:BIM}, a state-of-the-art method for shape matching. We test our approach under two graphs $\set{G}$. The first graph is a clique graph. The second graph connects each shape with $k$-nearest neighbor with respect to the GMDS descriptor~\cite{Rustamov:2007:LED} ($k=10$ in our experiments). 

\noindent\textbf{Evaluation setup.} We compare our approach to five baseline approaches, including three state-of-the-art approaches and two variants of our approach. Three state-of-the-art approaches are 1) functional-map based low-rank matrix recovery~\cite{DBLP:journals/tog/HuangWG14}, 2) point-map based low-rank matrix recovery via alternating minimization~\cite{DBLP:conf/iccv/ZhouZD15}, and 3) consistent partial matching via sparse modeling~\cite{DBLP:journals/cgf/CosmoRAMC17}. Two variants are 4) using a set of randomly sampled cycles~\cite{conf/cvpr/ZachKP10} whose size is the same as $|\set{B}|$, and 5) using the path-invariance basis derived from the fundamental cycle-basis of $\set{G}$ (c.f.~\cite{Kavitha:2009:SCB}) (which may contain long cycles). 

\begin{figure}
\centering
\includegraphics[width=0.49\columnwidth]{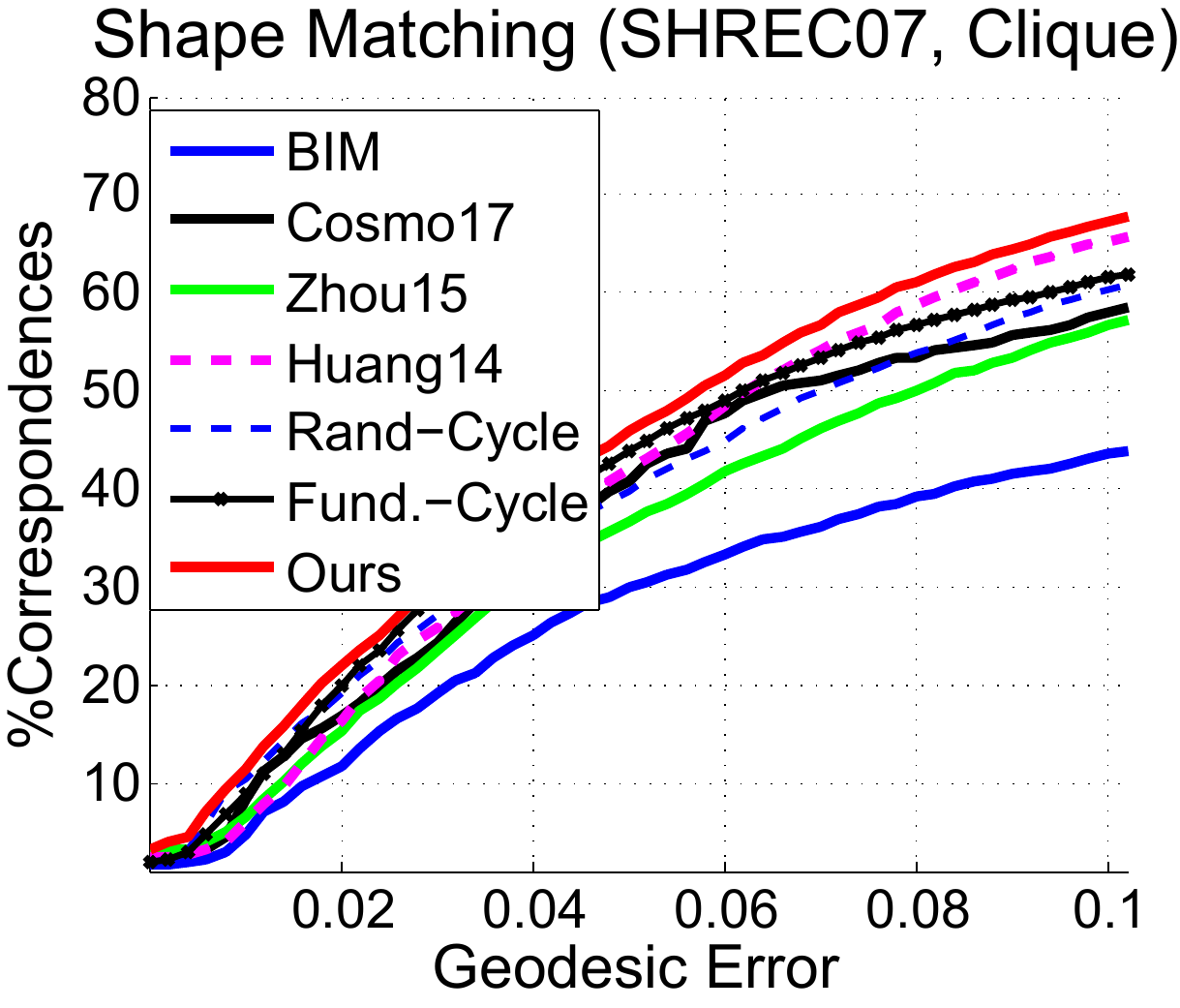}
\includegraphics[width=0.49\columnwidth]{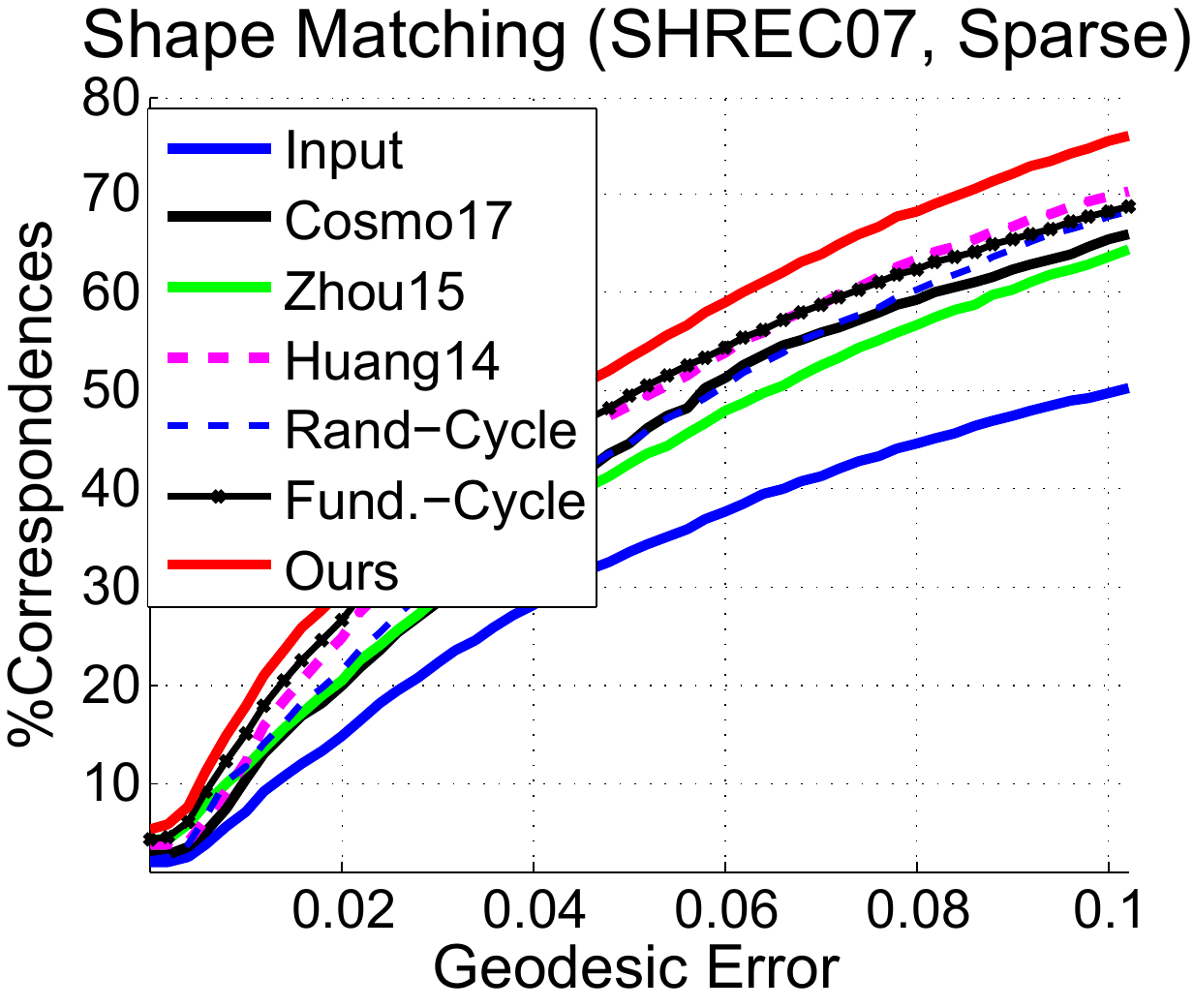}
\includegraphics[width=0.49\columnwidth]{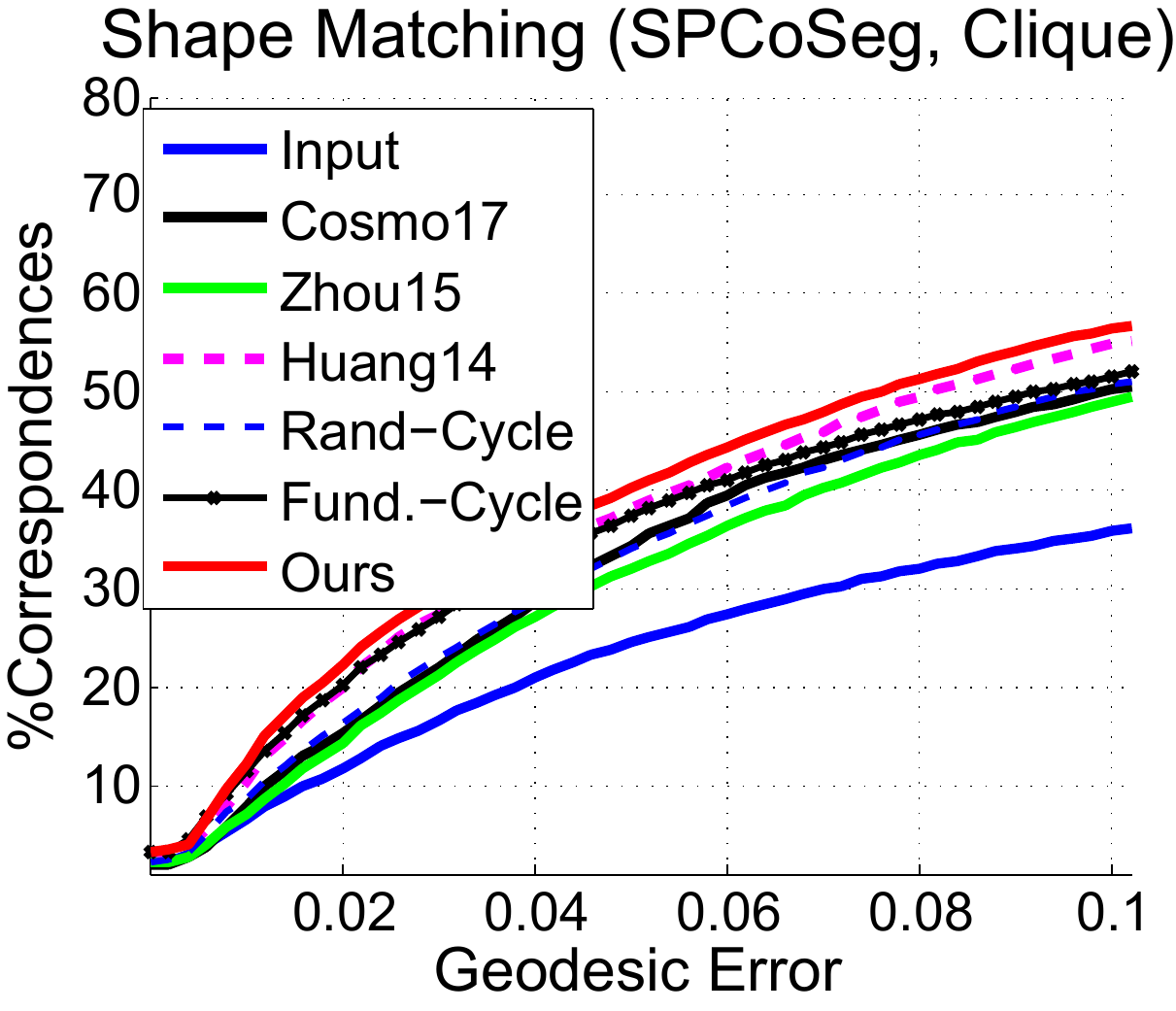}
\includegraphics[width=0.49\columnwidth]{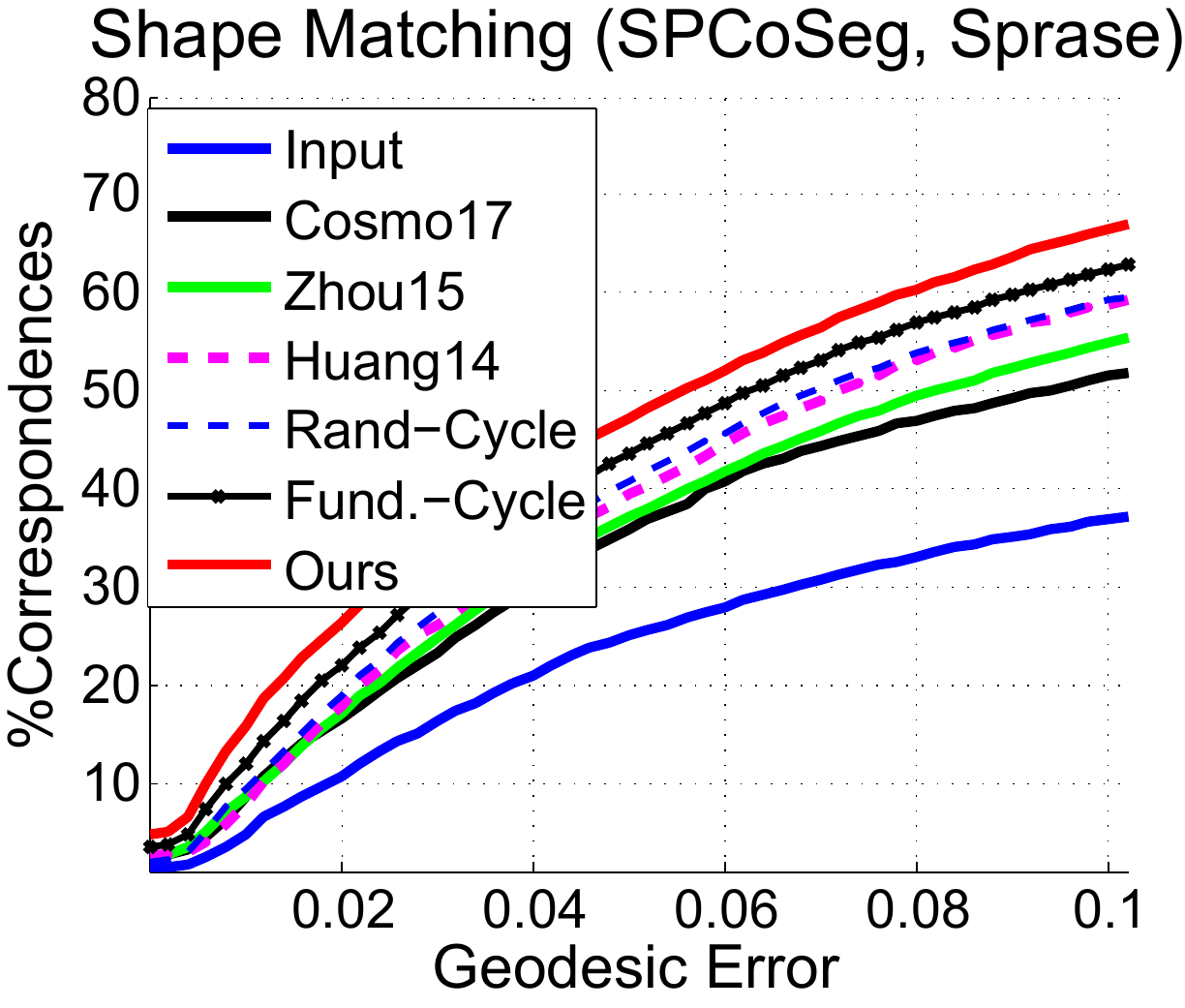}
\caption{\small{Baseline comparison on benchmark datasets. We show cumulative distribution functions (or CDFs) of each method with respect to annotated feature correspondences.}}
\label{Figure:CDF:Shape:Image}
\vspace{-0.25in}
\end{figure}

\begin{figure*}
\footnotesize{
\setlength\tabcolsep{3.33pt}
\begin{tabular}{c|ccccccccccccc|c}
& aero & bike & boat & bottle &bus &car & chair & table & mbike & sofa & train & tv & mean & \multirow{10}{*}{
 \includegraphics[width=0.285\textwidth]{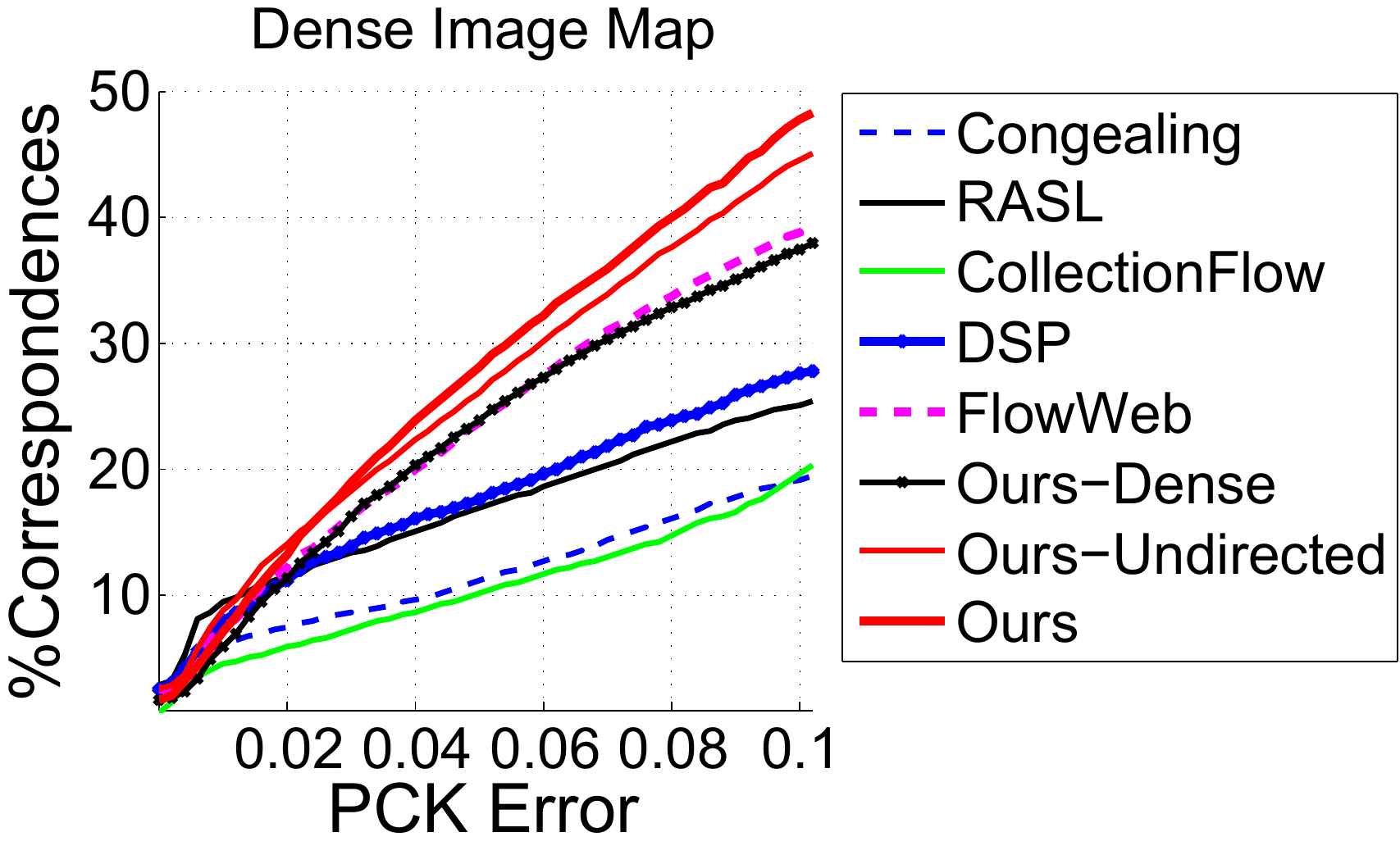}} \\
\cr Congealing &0.13 & 0.24 & 0.05 & 0.21 & 0.22 & 0.11 & 0.09
&0.05 &0.14 &0.09 &0.10 &0.09 &0.13& \\
RASL & 0.18 & 0.20 & 0.05 & 0.36 & 0.33 & 0.19 & 0.14 & 0.06 & 0.19 & 0.13 &0.14 & 0.29 & 0.19& \\
CollectionFlow & 0.17 & 0.18 & 0.06 & 0.33 & 0.31 & 0.15 & 0.15 & 0.04 & 0.12 & 0.11 & 0.10 & 0.11 & 0.12& \\
DSP & 0.19 & 0.33 & 0.07 & 0.21 & 0.36 & 0.37 & 0.12 & \textbf{0.07} & 0.19 & 0.13 & 0.15 & 0.21 & 0.20& \\
FlowWeb & 0.31 & 0.42 & \textbf{0.08} & 0.37 & 0.56 & 0.51 & 0.12 & 0.06 & 0.23 & 0.18 & 0.19 & 0.34 & 0.28& \\
Ours-Dense       & 0.29 & 0.42 & 0.07 & 0.39 & 0.53 & 0.55 & 0.11 & 0.06 & 0.22 & 0.18 & 0.21 & 0.31 & 0.28& \\
Ours-Undirected  & 0.32 & 0.43& 0.07 & 0.43 & 0.56 & 0.55& 0.18 & 0.06 & 0.26 & 0.21& \textbf{0.25}& 0.37 & 0.31& \\
Ours    & \textbf{0.35} & \textbf{0.45} & 0.07 & \textbf{0.45} & \textbf{0.63} & \textbf{0.62} & \textbf{0.19} & 0.06 & \textbf{0.27} & \textbf{0.22} & 0.23 & \textbf{0.38} & \textbf{0.33}&\\
\end{tabular}}
\caption{\small{(Left) Keypoint matching accuracy (PCK) on 12 rigid PASCAL VOC categories ($\alpha = 0.05$). Higher is better. (Right) Plots of the mean PCK of each method with varying $\alpha$}}
\label{Figure:Dense:Image:Flow:Quantitative}
\vspace{-0.15in}
\end{figure*}
\begin{figure*}
\footnotesize{
\setlength\tabcolsep{0.8pt}
\begin{tabular}{cccccccc} \hline
Source & Target & Congealing & RASL & CollectionFlow & DSP & FlowWeb & Ours \\
\includegraphics[width=0.122\textwidth]{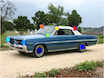} &
\includegraphics[width=0.122\textwidth]{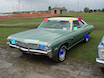} &
\includegraphics[width=0.122\textwidth]{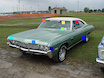} &
\includegraphics[width=0.122\textwidth]{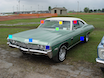} &
\includegraphics[width=0.122\textwidth]{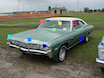} &
\includegraphics[width=0.122\textwidth]{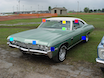} &
\includegraphics[width=0.122\textwidth]{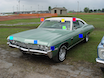} &
\includegraphics[width=0.122\textwidth]{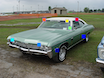} \\
\includegraphics[width=0.122\textwidth]{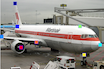} &
\includegraphics[width=0.122\textwidth]{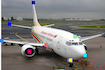} &
\includegraphics[width=0.122\textwidth]{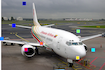} &
\includegraphics[width=0.122\textwidth]{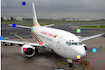} &
\includegraphics[width=0.122\textwidth]{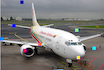} &
\includegraphics[width=0.122\textwidth]{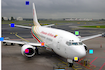} &
\includegraphics[width=0.122\textwidth]{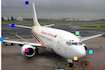} &
\includegraphics[width=0.122\textwidth]{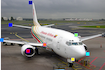} 
\end{tabular}}
\caption{\small{Visual comparison between our approach and state-of-the-art approaches. This figure is best viewed in color, zoomed in. More examples are included in the supplemental material.}}
\vspace{-0.15in}
\label{Figure:Dense:Image:Flow:Qualitative}
\end{figure*}

We evaluate the quality of each map through annotated key points (Please refer to the supplementary material). Following~\cite{Kim:2011:BIM,Huang:2013:CSM,DBLP:journals/tog/HuangWG14}, we report the cumulative distribution function (or CDF) of geodesic errors of predicted feature correspondences.

\noindent\textbf{Analysis of results.} Figure~\ref{Figure:CDF:Shape:Image} shows CDFs of our approach and baseline approaches. All participating methods exhibit considerable improvements from the initial maps, demonstrating the benefits of joint matching. Compared to state-of-the-art approaches, our approach is comparable when $\set{G}$ is a clique and exhibits certain performance gains when $\set{G}$ is sparse. One explanation is that low-rank approaches are based on relaxations of the cycle-consistency constraint (c.f.~\cite{Huang:2013:CSM}), and such relaxations become loose on sparse graphs. 
Compared to the two variants, our approach delivers the best results on both clique graphs and knn-graphs. This is because the two alternative strategies generate many long paths and cycles in $\set{B}$, making the total objective function (\ref{Eq:Total:Obj}) hard to optimize. On knn-graphs, both our approach and the baseline of using the fundamental cycle-basis outperform the baseline of randomly sampling path pairs, showing the importance of computing a path-invariance basis for enforcing the consistency constraint.

\subsection{Map Network of Dense Image Maps}
\label{Section:Network:Dense:Image:Flows}

In the second setting, we consider the task of optimizing dense image flows across a collection of relevant images. We again model this task using a map network $\set{F}$, where each domain $\set{D}_i$ is given by an image $I_i$. Our goal is to compute a dense image map $f_{ij}: I_i \rightarrow I_j$ (its difference to the identity map gives a dense image flow) between each pair of input images. To this end, we precompute initial dense maps $f_{ij}^{\init}, \forall (i,j)\in \set{E}$ using DSP~\cite{DBLP:conf/cvpr/KimLSG13}, which is a state-of-the-art approach for dense image flows. Our goal is to obtain improved dense image maps $f_{ij}, \forall (i,j)\in \set{E}$, which lead to dense image maps between all pairs of images in $\set{F}$ via map composition (See (\ref{Def:PathMap})). Due to scalability issues, state-of-the-art approaches for this task~\cite{Learned-Miller:2006:DDI,kemelmacher2012collection,Peng:2012:RRA,DBLP:conf/cvpr/ZhouLYE15} are limited to a small number of images. To address this issue, we encode dense image maps using the neural network $f^{\theta}$ described in~\cite{DBLP:conf/cvpr/ZhouKAHE16}. Given a fixed map network $\set{F}$ and the initial dense maps $f_{ij}^{\init},(i,j)\in \set{E}$, we formulate a similar optimization problem as (\ref{Eq:FMap:Opt}) to learn $\theta$:
\begin{equation}
\min\limits_{\theta} \quad 
\sum\limits_{(i,j)\in \set{E}} \|f^{\theta}_{ij}-f_{ij}^{\init}\|_1 + \lambda \sum\limits_{(p,q)\in \set{B}} \|f_{p}^{\theta}-f_{q}^{\theta}\|_{\set{F}}^2
\label{Eq:FMap:Opt3}
\end{equation}
where $\set{B}$ denotes a path-invariance basis associated with $\set{F}$; $p_s$ is the index of the start vertex of $p$; $f_{p}^{\theta}$ is the composite network along path $p$. 

\noindent\textbf{Dataset.} The image sets we use are sampled from 12 rigid categories of the PASCAL-Part dataset~\cite{DBLP:journals/corr/ChenMLFUY14}. To generate image sets that are meaningful to align, we pick the most popular view for each category (who has the smallest variance among 20-nearest neighbors). We then generate an image set for that category by collecting all images whose poses are within $30^{\circ}$ of this view. We construct the map network by connecting each image with 20-nearest neighbors with respect to the DSP matching score~\cite{DBLP:conf/cvpr/KimLSG13}. Note that the resulting $\set{F}$ is a directed graph as DSP is directed. The longest path varies between 4(Car)-6(Boat) in our experiments.

\noindent\textbf{Evaluation setup.} We compare our approach with Congealing~\cite{Learned-Miller:2006:DDI}, Collection Flow~\cite{kemelmacher2012collection}, RASL~\cite{Peng:2012:RRA}, and FlowWeb~\cite{DBLP:conf/cvpr/ZhouLYE15}.  
Note that both Flowweb and our approach use DSP as input. We also compare our approach against~\cite{DBLP:conf/cvpr/ZhouKAHE16} under a different setup (See supplementary material). To run baseline approaches, we follow the protocol of~\cite{DBLP:conf/cvpr/ZhouLYE15} to further break each dataset into smaller ones with maximum size of 100. In addition, we consider two variants of our approach: Ours-Dense and Ours-Undirected. Ours-Dense uses the clique graph for $\set{F}$. Ours-Undirected uses an undirected knn-graph, where each edge weight averages the bi-directional DSP matching scores (c.f.~\cite{DBLP:conf/cvpr/KimLSG13}). We employ the PCK measure~\cite{Yang:2013:AHD}, which reports the percentage of keypoints whose prediction errors fall within $\alpha\cdot \max(h,w)$ ($h$ and $w$ are image height and width respectively). 

\noindent\textbf{Analysis of results.} As shown in Figure~\ref{Figure:Dense:Image:Flow:Quantitative} and Figure~\ref{Figure:Dense:Image:Flow:Qualitative}, our approach outperforms all existing approaches across most of the categories. Several factors contribute to such improvements. First, our approach can jointly optimize more images than baseline approaches and thus benefits more from the data-driven effect of joint matching~\cite{Huang:2013:CSM,DBLP:conf/icml/ChenGH14}. This explains why all variants of our approach are either comparable or superior to baseline approaches. Second, our approach avoids fitting a neural network directly to dissimilar images and focuses on relatively similar images (other maps are generated by map composition), leading to additional performance gains. In fact, all existing approaches, which operate on sub-groups of similar images, also implicitly benefit from map composition. This explains why FlowWeb exhibits competing performance against Ours-Dense. Finally, Ours-Directed is superior to Ours-Undirected. This is because the outlier-ratio of $f_{ij}^{\init}$ in Ours-Undirected is higher than that of Ours-Directed, which selects edges purely based on matching scores. 

\begin{figure*}
\centering
\setlength\tabcolsep{2pt}
\begin{tabular}{l | l l l l}
Ground Truth & 8\% Label & 30\% Label & 100\% Label & 8\% Label + 92\%Unlabel \\\hline
\multirow{2}{*}{
\includegraphics[width=0.4\columnwidth]{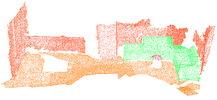}}
&
\multicolumn{1}{l}{\includegraphics[width=0.4\columnwidth]{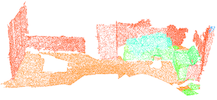}}
&
\multicolumn{1}{l}{\includegraphics[width=0.4\columnwidth]{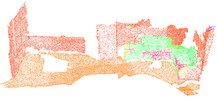}}
&
\multicolumn{1}{l}{\includegraphics[width=0.4\columnwidth]{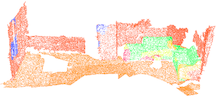}}
&
\multicolumn{1}{l}{\includegraphics[width=0.4\columnwidth]{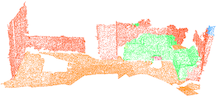}}\\\cline{2-5}
&
\multicolumn{1}{l}{\includegraphics[width=0.4\columnwidth]{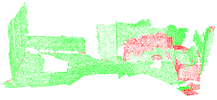}}
&
\multicolumn{1}{l}{\includegraphics[width=0.4\columnwidth]{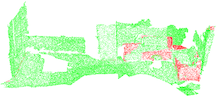}}
&
\multicolumn{1}{l}{\includegraphics[width=0.4\columnwidth]{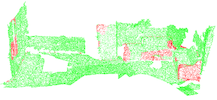}}
&
\multicolumn{1}{l}{\includegraphics[width=0.4\columnwidth]{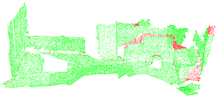}}\\\hline
\multirow{2}{*}{
\includegraphics[width=0.4\columnwidth]{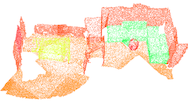}}
&
\multicolumn{1}{l}{\includegraphics[width=0.4\columnwidth]{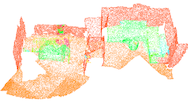}}
&
\multicolumn{1}{l}{\includegraphics[width=0.4\columnwidth]{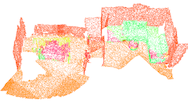}}
&
\multicolumn{1}{l}{\includegraphics[width=0.4\columnwidth]{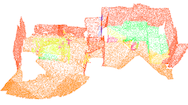}}
&
\multicolumn{1}{l}{\includegraphics[width=0.4\columnwidth]{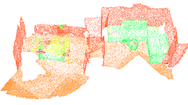}}\\\cline{2-5}
&
\multicolumn{1}{l}{\includegraphics[width=0.4\columnwidth]{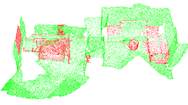}}
&
\multicolumn{1}{l}{\includegraphics[width=0.4\columnwidth]{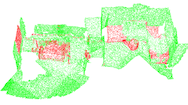}}
&
\multicolumn{1}{l}{\includegraphics[width=0.4\columnwidth]{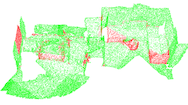}}
&
\multicolumn{1}{l}{\includegraphics[width=0.4\columnwidth]{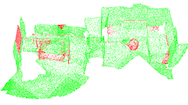}}\\\hline
\multirow{2}{*}{
\includegraphics[width=0.4\columnwidth]{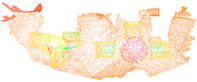}}
&
\multicolumn{1}{l}{\includegraphics[width=0.4\columnwidth]{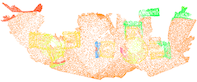}}
&
\multicolumn{1}{l}{\includegraphics[width=0.4\columnwidth]{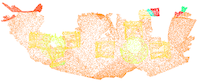}}&
\multicolumn{1}{l}{\includegraphics[width=0.4\columnwidth]{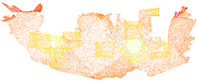}}
&
\multicolumn{1}{l}{\includegraphics[width=0.4\columnwidth]{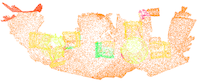}}\\\cline{2-5}
&
\multicolumn{1}{l}{\includegraphics[width=0.4\columnwidth]{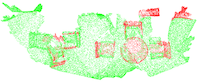}}
&
\multicolumn{1}{l}{\includegraphics[width=0.4\columnwidth]{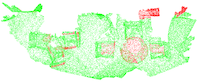}}
&
\multicolumn{1}{l}{\includegraphics[width=0.4\columnwidth]{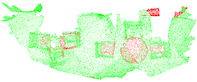}}&
\multicolumn{1}{l}{\includegraphics[width=0.4\columnwidth]{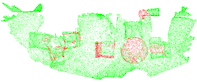}}\\\hline
\end{tabular}
\caption{\small{Qualitative comparisons of 3D semantic segmentation results on ScanNet~\protect\cite{dai2017scannet}.} Each row represents one testing instance, where ground truth and top sub-row show prediction for 21 classes and bottom sub-row only shows correctly labeled points. (Green indicates correct predictions, while red indicates false predictions.) This figure is best viewed in color, zoomed in. }
\label{Figure:Map:Network:3D:Understanding}
\vspace{-0.2in}
\end{figure*}

\subsection{Map Network of 3D Representations}
\label{Section:Network:3D:Representation}

In the third setting, we seek to jointly optimize a network of neural networks to improve the performance of individual networks. We are particularly interested in the task of semantic segmentation of 3D scenes. Specifically, we consider a network with seven 3D representations (See Figure~\ref{Figure:Best:Result}). 
The first representation is the input mesh. The last representation is the space of 3D semantic segmentations. The second to fourth 3D representations are point clouds with different number of points: PCI (12K), PCII (8K), and PCIII(4K). The motivation of varying the number of points is that the patterns learned under different number of points show certain variations, which are beneficial to each other. In a similar fashion, the fifth and sixth are volumetric representations under two resolutions: VOLI($32\times 32\times 32$) and VOLII($24\times 24\times 24$). The directed maps between different 3D representations fall into three categories, which are summarized below:

\noindent\textsl{1. Segmentation networks.} We use  PointNet++~\cite{DBLP:journals/corr/QiYSG17} and 3D U-Net\cite{cciccek20163d} for the segmentation networks under point cloud and volumetric representations, respectively. 

\noindent\textsl{2. Pointcloud sub-sampling maps.} We have six pointcloud sub-sampling maps among the mesh representation (we uniformly sample 24K points using~\cite{Osada:2002:SD}) and three point cloud representations. For each point sub-sampling map, we force the down-sampled point cloud to align with the feature points of the input point cloud~\cite{10.1111:1467-8659.00675}. Note that this down-sampled point cloud is also optimized through a segmentation network to maximize the segmentation accuracy.   

\noindent\textsl{3. Generating volumetric representations.} Each volumetric representation is given by the signed-distance field (or SDF) described in \cite{song2016ssc}. These SDFs are precomputed. 

\begin{table}[t]
    \small{
    \setlength\tabcolsep{1.6pt}
    \centering
    \begin{tabular}{l|cccccc}
    & PCI & PCII & PCIII & VOLI & VOLII & ENS \\ \hline
    100\% Label (Isolated) & 84.2 & 83.3 & 83.4 & 81.9 & 81.5 & 85 \\
    8\% Label (Isolated)  & 79.2 & 78.3 & 78.4 & 78.7 & 77.4 & 81.4\\
    8\% Label + Unlabel (Joint) & 82.3 & 82.5 & 82.3 & 81.6 & 79.0 & 83.4\\
    30\% Label (Isolated) & 80.8 & 81.9 & 81.2 & 80.3 & 79.5 & 83.2
    \end{tabular}}
    \caption{\small{Semantic surface voxel label prediction accuracy on ScanNet test scenes (in percentages), following ~\cite{DBLP:journals/corr/QiYSG17}. We also show the ensembled prediction accuracy with five representations in the last column. }}
    \label{Figure:Map:Network:3D:Result}
    \vspace{-0.2in}
\end{table}
\noindent\textbf{Experimental setup.} We have evaluated our approach on ScanNet semantic segmentation benchmark~\cite{dai2017scannet}. Our goal is to evaluate the effectiveness of our approach when using a small labeled dataset and a large unlabeled dataset. To this end, we consider three baseline approaches, which train the segmentation network under each individual representation using 100\%, 30\%, and 8\% of the labeled data. 
We then test our approach by utilizing 8\% of the labeled data, which defines the data term in (\ref{Eq:Total:Obj}), and 92\% of the unlabeled data, which defines the regularization term of (\ref{Eq:Total:Obj}). We initialize the segmentation network for point clouds using uniformly sampled points trained on labeled data. We then fine-tune the entire network using both labeled and unlabeled data. Note that unlike \cite{DBLP:journals/corr/abs-1804-08328}, our approach leverages essentially the same labeled data but under different 3D representations. The boost in performance comes from unlabeled data. Code is publicly available at \url{https://github.com/zaiweizhang/path\_invariance\_map\_network}. 


\noindent\textbf{Analysis of results.} Figure~\ref{Figure:Map:Network:3D:Understanding} and Table~\ref{Figure:Map:Network:3D:Result} present qualitative and quantitative comparisons between our approach and baselines. Across all 3D representations, our approach leads to consistent improvements, demonstrating the robustness of our approach. Specifically, when using 8\% labeled data and 92\% unlabeled data, our approach achieves competing performance as using 30\% to 100\% labeled data when trained on each individual representation. Moreover, the accuracy on VOLI is competitive against using 100\% of labeled data, indicating that the patterns learned under the point cloud representations are propagated to train the volumetric representations. 
We also tested the performance of applying popular vote~\cite{Rokach:2010:EC} on the predictions of using different 3D representations. The relative performance gains remain similar (See the last column in Table\ref{Figure:Map:Network:3D:Result}). Please refer to Appendix~\ref{Section:Detail:Network:3D:Understanding} for more experimental evaluations and baseline comparisons.
\section{Conclusions}
\label{Section:Conclusions}

We have studied the problem of optimizing a directed map network while enforcing the path-invariance constraint via path-invariance bases. We have described an algorithm for computing a path-invariance basis with polynomial time and space complexities. The effectiveness of this approach is demonstrated on three groups of map networks with diverse applications.

\noindent\textbf{Acknowledgement.} Qixing Huang would like to acknowledge support from NSF DMS-1700234, NSF CIP-1729486, NSF IIS-1618648, a gift from Snap Research and a GPU donation from Nvidia Inc. Xiaowei Zhou is supported in part by NSFC (No. 61806176) and Fundamental Research Funds for the Central Universities.



\newpage
\clearpage
\appendix

\section{Proof of Theorems and Propositions}
\label{Section:proof}

\subsection{Proof of Proposition 1}
To show that $\set{F}$ is path-invariant, it suffices to prove that $f_p=f_q$ for every path pair $(p,q)\in \set{G}_{\pair}$. But by Definition 7, $(p,q)$ is either in $\set{G}_{\pair}$ or can be induced from a finite number of operations with \textup{merge}, \textup{stitch} and/or \textup{cut}. So if we can show that the output path pair in every round of operation keeps consistency on the map network $\set{F}$, given the input path pairs are consistent, then all path pairs on $\set{G}$ would be path-invariant by employing an induction proof. Next we achieve this goal by considering three operations respectively.

\begin{itemize}
    \item \textbf{merge.} The merge operation takes as input two path pairs $(p,q),(p',q')\in\set{G}_\pair$ where $p'=r\sim p\sim r'$, i.e., $p'$ is formed by stitching three sub-paths $r,p$ and $r'$ in order. By Definition 2, it is easy to see that
    $$
    f_{p'}=f_{r'}\circ f_{p}\circ f_r.
    $$
    But we are given that $\set{F}$ is consistent on the input pairs, or equivalently,
    $$
    f_{p}=f_{q},\quad f_{p'}=f_{q'}.
    $$
    Hence
    $$
    f_{q'}=f_{p'}=f_{r'}\circ f_{q}\circ f_{r}=f_{r\sim q\sim r'}.
    $$
    So $\set{F}$ is also consistent on path pair $(r\sim q\sim r',q')$.
    
    \item \textbf{stitch.} The stitch operation takes as input two path pairs $(p,q),(p',q')$ where $p,q\in\set{G}_{\path}(u,v)$ and $p',q'\in\set{G}_{\path}(v,w)$. Since $\set{F}$ is consistent on $(p,q)$ and $(p',q')$, it follows immediately
    $$
    f_{p\sim p'}=f_{p'}\circ f_{p}=f_{q'}\circ f_{q}=f_{q\sim q'},
    $$
    which means $\set{F}$ is also consistent on $(p\sim p',q\sim q')$.
    
    \item \textbf{cut.} The cut operation takes as input two path pairs $(\set{C}_1,\emptyset)$ and $(\set{C}_2,\emptyset)$, where $\set{C}_1$ and $\set{C}_2$ are two common vertices $u,v$ and share a common intermediate path from $v$ to $u$. The two cycles can be represented by $u\xrightarrow{p}v\xrightarrow{q}u$ and $u\xrightarrow{p'}v\xrightarrow{q}u$ where $p,p'\in\set{G}_\path(u,v)$ and $q\in\set{G}_\path(v,u)$. Since $\set{F}$ is consistent on $(\set{C}_1,\emptyset)$ and $(\set{C}_2,\emptyset)$, we have
    $$
    f_{p\sim q}=f_q\circ f_p=I,\quad f_{p'\sim q}=f_{q}\circ f_{p'}=I.
    $$
    However, it is known that the inverse of some function must be unique, giving the following result
    $$
    f_{p'}=f_{p},
    $$
    or in other words, $\set{F}$ is consistent on path pair $(p,p')$.
    
\end{itemize}
    
    The consistency of $\set{F}$ on the output pairs for all three operations given the consistency on their input pairs ensures our proposition.\qed

\subsection{Proof of Theorem 3.1}
The algorithm adds exactly $|\set{E}|$ edges in total. And during each edge insertion, at most $|\set{P}|\leq |\set{V}|$ path pairs would be added to $\set{B}_{dag}(\sigma)$, thus it follows immediately that $|\set{B}_{dag}(\sigma)|\leq |\set{V}||\set{E}|$.

Next we show that $\set{B}_{dag}(\sigma)$ indeed is a path-invariance basis for $\set{G}$. To this end, we will verify that every path pair in $\set{G}_{\pair}$ can be induced from a subset of $\set{B}_{dag}(\sigma)$ by operations, using a induction proof. In particular, we claim that at all time points, all path pairs in $\set{G}_{cur}$ can be induced from $\set{B}_{dag}(\sigma)$ by a series of operations. Initially, this inductive assumption holds trivially since $\set{G}_{cur}$ is an empty set.

\captionsetup{font=small}
\begin{wrapfigure}{R}{0.3\columnwidth}
\centering
\includegraphics[width=0.3\columnwidth]{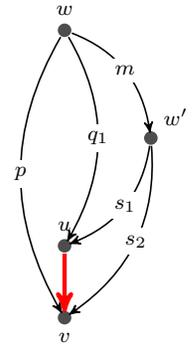}
\caption{An Illustration for Path-Pair Generation}
\end{wrapfigure}

Suppose now we were processing an edge $(u,v)\in\set{E}$ (so $(u,v)\not\in \set{G}_{cur}(\sigma)$ at this time point) and let $\set{G}_{cur}'=\set{G}_{cur}\cup\{(u,v)\}$. By inductive assumption, all path pairs in $\set{G}_{cur}$ can be induced from $\set{B}_{dag}(\sigma)$. After inserting $(u,v)$ into $\set{G}_{cur}$, it suffices to consider path pairs that contain edge $(u,v)$ since all other path pairs have been guaranteed by inductive assumption. Let $(p,q)$ be a path pair in $\set{G}_{cur}'$ containing $(u,v)$. Without loss of generality, suppose $p,q\in\set{G}_{\path}(w,v)$ and $$p=p_1\sim (r,v),\ q=q_1\sim (u,v).$$ If $r=u$, then $(p,q)$ can be induced by stitching $(p_1,q_1)$ and $((u,v),(u,v))$ where $(p_1,q_1)\in\set{G}_{cur}$. We assume $r\neq u$, and then $p$ would be a path from $w$ to $v$ in $\set{G}_{cur}$ and $q_1$ would be a path from $w$ to $u$ in $\set{G}_{cur}$.

Recall the definition of $\set{P}$ and $\overline{\set{P}}$. $w\in\set{P}$ immediately follows.  If $w\not\in \overline{\set{P}}$, then there exists $w'\neq w$ such that $w$ can reach $w'$ in $\set{G}_{cur}$ and $w'\in\overline{\set{P}}$ and denote such path as $m$. For convenience, we let $w'=w$ and $m=\emptyset$ when $w\in \overline{\set{P}}$. Every vertex in $\overline{\set{P}}$ corresponds to a path-invariance pair to be added to $\set{B}_{dag}$ by our algorithm. Here we assume that it is $(s_1\sim (u,v),s_2)$ for $w'$ where $s_1\in\set{G}_{\path}(w',u)$, $s_2\in\set{G}_{\path}(w',v)$ and $s_2$ is within $\set{G}_{cur}$.

By the property of DAG and the order of edge insertion, all paths from $w$ to $u$ in $\set{G}$ are also in $\set{G}_{cur}$ since $(u,v)\in\set{E}$. Thus $(q_1,m\sim s_1)$ can be induced from $\set{B}_{dag}(\sigma)$ by inductive assumption. Similarly, as $s_2$ is within $\set{G}_{cur}$, $(p,m\sim s_2)$ is also a path-invariance pair, which can be induced from $\set{B}_{dag}(\sigma)$. Next we give the operation steps to build $(p,q)$:
\begin{align}
    &(m,m)+(s_1\sim (u,v),s_2)\nonumber\\
    &\hspace{1cm}\xrightarrow{stitch} (m\sim s_1\sim (u,v),m\sim s_2)\\
    &(m\sim s_1\sim (u,v),m\sim s_2) + (q_1,m\sim s_1)\nonumber\\
    &\hspace{1cm}\xrightarrow{merge} (q_1\sim (u,v),m\sim s_2)\\
    &(q_1\sim (u,v),m\sim s_2) + (p,m\sim s_2)\nonumber\\
    &\hspace{1cm}\xrightarrow{merge} (p,q)
\end{align}
For the last step, notice that $q=q_1\sim (u,v)$ and $(p,q)$ is equivalent to $(q,p)$. Thus all path pairs in $\set{G}_{cur}'$ can be induced by path pairs in $\set{B}_{dag}(\sigma)$ with a series of operations, which completes our proof by induction.
\qed

\subsection{Proof of Proposition 2}
\label{Subsection:Proof:Proposition:2}

Before proving Proposition 2, we first introduce some well-known terms for depth-first search. There are two time stamps $d[v]$ and $f[v]$ for each vertex $v$, where $d$ is defined as the time point when it visits $v$ for the first time and $f$ as the time point when it finishes visiting $v$. Some edge $(u,v)$ in $\set{E}$ can be classified into one of four disjoint types as follows:
\begin{itemize}
    \item \textbf{Tree Edge:} $v$ is visited for the first time as we traverse the edge $(u,v)$. In this case $(u,v)$ will be added into the resulting DFS spanning tree. For tree edge we have
    $$d[u]<d[v],\quad f[u]>f[v].$$
    \item \textbf{Back Edge:} $v$ is visited and is an ancestor of $u$ in the current spanning tree. For back edge we have $$d[u]>d[v],\quad f[u]<f[v].$$
    \item \textbf{Forward Edge:} $u$ is visited and is an ancestor of $v$ in the current spanning tree. For forward edge we have
    $$d[u]<d[v],\quad f[u]>f[v].$$
    \item \textbf{Cross Edge:} $v$ is visited and is neither an ancestor nor descendant of $u$ in the current spanning tree. For cross edge we have
    $$d[u]>d[v],\quad f[u]>f[v].$$
\end{itemize}
Using these definitions, we prove that:

\textit{Any cycle $\set{C}$ in $\set{G}$ have a vertex $u$ in $\set{C}$ such that all other vertices are located within the sub-tree rooted at $u$, i.e., are the descendants of $u$ in $\set{T}$.}

Let
$$\set{C}: u_1\to u_2\to\dots\to u_k\to u_1.$$
Without loss of generality, $u_1$ is assumed to be the one with smallest $d$ among all $\{u_i\}$. If not all $u_i$ are descendants of $u_1$, we choose $u_t$ to be the one with smallest $t$, which means $u_{t-1}$ is a descendant of $u_1$ but $u_t$ is not. Obviously $(u_{t-1},u_t)$ cannot be a tree edge or forward edge, which causes $u_t$ to be a descendant of $u_{t-1}$ and also a descendant of $u_1$. If $(u_{t-1},u_t)$ is a back edge, then $u_t$ is not a descendant of $u_1$ if and only if $u_{t-1}=u_1$ since there is in fact unique back path in the spanning tree $\set{T}$. But $u_{t-1}=u_1$ means $u_t$ is a parent of $u_1$, and thus there exists a smaller $d$ than $u_1$, which results in a contradiction. Also $(u_{t-1},u_t)$ cannot be a cross edge. In fact, since $u_{t-1}$ is a descendant of $u_1$, we have $f[u_{t-1}]<f[u_1]$. Together with $f[u_t]<f[u_{t-1}]$ from cross edge property, we have $f[u_t]<f[u_1]$. But $u_t$ is not a descendant or ancestor of $u_1$, which means the sub-tree rooted at $u_1$ is disjoint from the sub-tree rooted at $u_t$, so intervals $[d[u_1],f[u_1]]$ and $[d[u_t],f[u_t]]$ must be disjoint by the property of depth-first search. As thus $f[u_t]<f[u_1]$ implies $d[u_t]<d[u_1]$, which contradicts the assumption that $d[u_1]$ is smallest among $u_i$. Hence all $u_i$ are descendants of $u_1$.

Now come back to the original proposition. Continue using the notation $\set{C}$ defined above. In addition we define $\set{C}_i$ as the sub-path from $u_1$ to $u_i$, i.e.,
$$
\set{C}_i: u_1\to u_2\to\dots\to u_i.
$$

We will show $\set{C}$ can be induced from $\set{B}$ by a finite number of operations with merge, stitch and cut. Above all, we have assumed the property of path-invariance on $\set{T}$ by Theorem 3.1. Given $u_1$ is the common ancestor of all $u_i$, we inductively prove the following statement:

\textit{The path $u_1\to u_2\to\dots\to u_t$ $(t\leq k)$ is equivalent to the tree path from $u_1$ to $u_t$. Here tree path means a path in which all edges are in the spanning tree $\set{T}$.}

The base case is trivial. Now suppose $u_1\to u_2\to\dots \to u_t$ $(t<k)$ is equivalent to tree path $P$ from $u_1$ to $u_t$ and we continue to check $u_1\to u_2\to\dots\to u_{t+1}$.

\begin{itemize}
\item If $(u_t,u_{t+1})$ is a tree edge, then $P\sim(u_t,u_{t+1})$ is still a tree path and a stitch operation on path pair $(\set{C}_t,P)$ and $((u_t,u_{t+1}),(u_t,u_{t+1}))$ gives the equivalency that we want.

\item If $(u_t,u_{t+1})$ is a forward edge, then there exists a tree path $P_1$ from $u_t$ to $u_{t+1}$. By path-invariance on $\set{G}_{dag}$, we can stitch two path-invariance pair $(\set{C}_t,P)$ and $((u_{t},u_{t+1}),P_1)$ to obtain the desired equivalency.

\item If $(u_t,u_{t+1})$ is a back edge, then there exists a tree path $P_1$ from $u_{t+1}$ to $u_{t}$. In addition by our construction the cycle $P_1\sim(u_t,u_{t+1})$ has been added into our basis set $\set{B}$. Denote the tree path from $u_1$ to $u_{t+1}$ as $P_2$, then stitching $(P_2,P_2)$ and $(P_1\sim (u_t,u_{t+1}),\emptyset)$ gives $(P_2\sim P_1\sim (u_t,u_{t+1}),P_2)$. On the other hand, by inductive assumption we have path-invariance pair $(P_2\sim P_1,\set{C}_t)$ since $P_2\sim P_1$ is just the tree path from $u_1$ to $u_t$. Thus by merging $(P_2\sim P_1,\set{C}_t)$ and $(P_2\sim P_1\sim (u_t,u_{t+1}),P_2)$ we obtain the path pair $(\set{C}_t\sim (u_t,u_{t+1}),P_2)$, or equivalently, $(\set{C}_{t+1}, P_2)$.

\item If $(u_t,u_{t+1})$ is a cross edge, then $(u_t,u_{t+1})$ has been included in $\set{G}_{dag}$. Denote by $P_1$ the tree path from $u_1$ to $u_{t}$. In this way all $P_1\sim (u_t,u_{t+1})$ would be equivalent to another tree path $P_2$ from $u_1$ to $u_{t+1}$ since all edges involved here are within $\set{G}_{dag}$ which maintains all possible path-invariance pairs. By merging path pairs $(P_1\sim (u_t,u_{t+1}),P_2)$ and $(P_1,\set{C}_{t})$ we obtain path pair $(\set{C}_t\sim (u_t,u_{t+1}),P_2)$, or $(\set{C}_{t+1},P_2)$, which is exactly we want to verify.
\end{itemize}

As thus we finished our inductive proof. In particular, the path (also a cycle) $u_1\to\dots\to u_k\to u_1$ is equivalent to $\emptyset$, or more precisely, the path pair $(\set{C},\emptyset)$ can be induced from $\set{B}$ by a finite number of  merge and stitch operations.

To complete our proof, we need to show that all path pairs in $\set{G}$ instead of just $\set{G}_{dag}$ can be induced from $\set{B}$. This is relatively easy. Consider two path $P_1$ and $P_2$ both from $u$ to $v$. Since $\set{G}$ is strongly connected, there must exist some path $P_3$ from $v$ to $u$. The cut operation on $P_1\sim P_3$ and $P_2\sim P_3$ for the common vertices $u$ and $v$ immediately gives the path pair $(P_1,P_2)$.
\qed

\begin{figure*}
\footnotesize{
\setlength\tabcolsep{0.8pt}
\begin{tabular}{cccccccc} \hline
Source & Target & Congealing & RASL & CollectionFlow & DSP & FlowWeb & Ours \\
\includegraphics[width=0.122\textwidth]{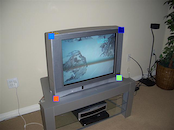} &
\includegraphics[width=0.122\textwidth]{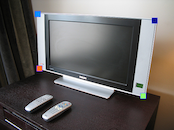} &
\includegraphics[width=0.122\textwidth]{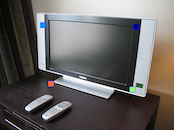} &
\includegraphics[width=0.122\textwidth]{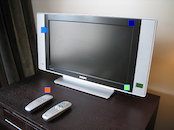} &
\includegraphics[width=0.122\textwidth]{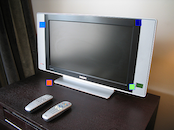} &
\includegraphics[width=0.122\textwidth]{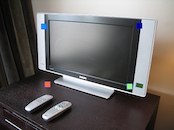} &
\includegraphics[width=0.122\textwidth]{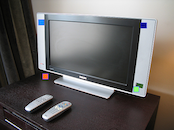} &
\includegraphics[width=0.122\textwidth]{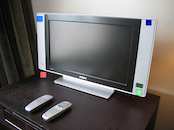} \\
\includegraphics[width=0.122\textwidth]{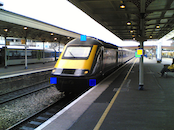} &
\includegraphics[width=0.122\textwidth]{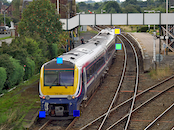} &
\includegraphics[width=0.122\textwidth]{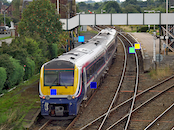} &
\includegraphics[width=0.122\textwidth]{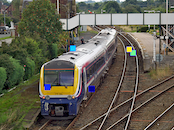} &
\includegraphics[width=0.122\textwidth]{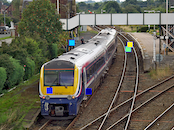} &
\includegraphics[width=0.122\textwidth]{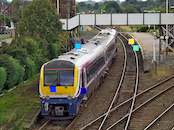} &
\includegraphics[width=0.122\textwidth]{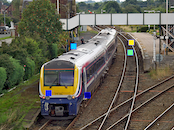} &
\includegraphics[width=0.122\textwidth]{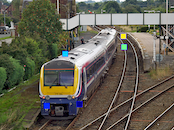} \\
\includegraphics[width=0.122\textwidth]{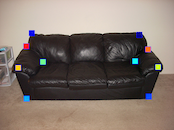} &
\includegraphics[width=0.122\textwidth]{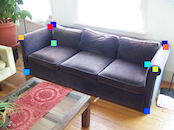} &
\includegraphics[width=0.122\textwidth]{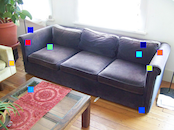} &
\includegraphics[width=0.122\textwidth]{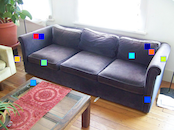} &
\includegraphics[width=0.122\textwidth]{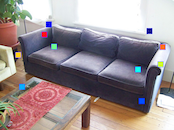} &
\includegraphics[width=0.122\textwidth]{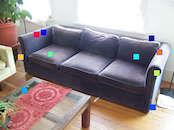} &
\includegraphics[width=0.122\textwidth]{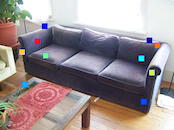} &
\includegraphics[width=0.122\textwidth]{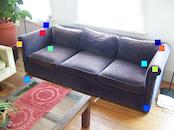} \\
\includegraphics[width=0.122\textwidth]{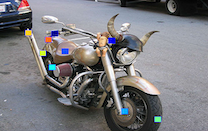} &
\includegraphics[width=0.122\textwidth]{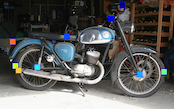} &
\includegraphics[width=0.122\textwidth]{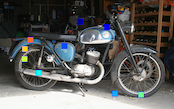} &
\includegraphics[width=0.122\textwidth]{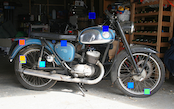} &
\includegraphics[width=0.122\textwidth]{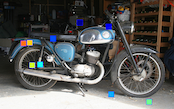} &
\includegraphics[width=0.122\textwidth]{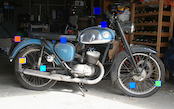} &
\includegraphics[width=0.122\textwidth]{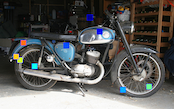} &
\includegraphics[width=0.122\textwidth]{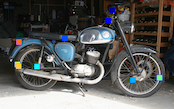} \\
\includegraphics[width=0.122\textwidth]{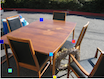} &
\includegraphics[width=0.122\textwidth]{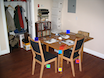} &
\includegraphics[width=0.122\textwidth]{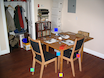} &
\includegraphics[width=0.122\textwidth]{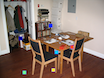} &
\includegraphics[width=0.122\textwidth]{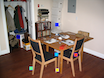} &
\includegraphics[width=0.122\textwidth]{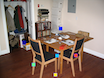} &
\includegraphics[width=0.122\textwidth]{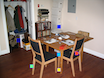} &
\includegraphics[width=0.122\textwidth]{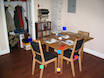} 
\end{tabular}}
\caption{\small{Visual comparison between our approach and state-of-the-art approaches. This figure is best viewed in color, zoomed in.}}
\vspace{-0.15in}
\label{Figure:Dense:Image:Flow:QualitativeI}
\end{figure*}

\subsection{Proof of Theorem 3.2}

To prove this theorem, we first prove the following lemma:

\noindent 
\textbf{Lemma}\textit{ Suppose $\set{G}_i$ and $\set{G}_j$ are two strongly connected components in $\set{G}$ with $(\set{G}_i,\set{G}_j)\in\set{G}_{dag}$. Given any vertices $u,u'\in \set{G}_i$ and $v,v'\in\set{G}_j$ with $(u,v),(u',v')\in\set{E}_{ij}$, and paths $p\in\set{G}_{\path}(u,u')$, $p'\in\set{G}_{\path}(v,v')$, we claim that $p\sim (u',v')$ is equivalent to $(u,v)\sim p'$ under $\set{B}$. 
}

In fact since $\set{B}$ ensures equivalence for all path pairs inside the same SCC, the specific $p,p'$ does not matter. We only care about the starting and ending points when everything happens inside a single SCC. So in the following proof we will use $P(x,y)$ to denote some path from $x$ to $y$ inside the single SCC but not mentioning the intermediate vertices. Recall that we built a (undirected) spanning tree $\set{T}$ on $\set{E}_{ij}^2$. Thus we have an edge sequence
$$
(u_1,v_1), (u_2,v_2),\dots, (u_k,v_k)
$$
where $u_1=u$, $v_1=v$, $u_k=u',v_k=v'$, and edge pair $$((u_l,v_l),(u_{l+1},v_{l+1}))$$
are in $\set{T}$ for all $l=1,\dots,k-1$. Next we inductively prove that $P(u_1,u_t)\sim (u_t,v_t)$ is equivalent to $(u_1,v_1)\sim P(v_1,v_t)$ for $t=1,\dots,k$. The base case where $t=1$ is trivial. Given the correctness for $t$, consider $t+1$. It is known that $(u_t,v_t)\sim P(v_t,v_{t+1})$ is equivalent to $P(u_t,u_{t+1})\sim (u_{t+1},v_{t+1})$ by the construction of $\set{T}$ and $P(u_1,u_t)\sim (u_t,v_t)$ is equivalent to $(u_1,v_1)\sim P(v_1,v_t)$ by inductive assumption. By successively applying two merge operations on path $$(u_1,v_1)\sim P(v_1,v_t)\sim P(v_t,v_{t+1})$$ we obtain the equivalent path
$$
P(u_1,u_t)\sim P(u_t,u_{t+1})\sim (u_{t+1},v_{t+1}).
$$
But it is straightforward that $P(u_1,u_t)\sim P(u_t,u_{t+1})$ is equivalent to $P(u_1,u_{t+1})$ under $\set{B}$ since $u_1,u_t,u_{t+1}$ are in the same SCC. Similarly, $P(v_1,v_t)\sim P(v_t,v_{t+1})$ is equivalent to $P(v_1,v_{t+1})$. Thus finally we obtain the equivalency on $P(u_1,u_{t+1})\sim (u_{t+1},v_{t+1})$ and $(u_1,v_1)\sim P(v_1,v_{t+1})$, which completes our inductive proof and the lemma immediately follows.

Come back to the original theorem. With notation $P(x,y)$, we can express an arbitrary path $p$ in $\set{G}$ from $u$ to $v$ as
\begin{align*}
p: P(u_1,v_1)&\sim (v_1,u_2)\sim P(u_2,v_2)\sim\\
&\quad\dots \sim (v_{k-1},u_k)\sim P(u_k,v_k)
\end{align*}
where $u_1=u$, $v_k=v$, and $u_i$, $v_i$ are in the same SCC $\set{G}_{b_i}$. Similarly write another path $p'$ from $u$ to $v$ this way:
\begin{align*}
p': P(u_1',v_1')&\sim (v_1',u_2')\sim P(u_2',v_2')\sim\\
&\quad\dots \sim (v_{k-1}',u_k')\sim P(u_k',v_k')
\end{align*}
where $u_1'=u$, $v_k'=v$, and $u_i$, $v_i$ are in the same SCC $\set{G}_{b_i'}$ with obvious constraints $b_1=b_1'$ and $b_k=b_k'$. As we extend $\set{B}_{dag}$ that maintains the equivalency on all possible pairs in $\set{G}_{dag}$ to $\set{G}$, there would be a path pair
\begin{align*}
q: P(\alpha_1,\beta_1)&\sim (\beta_1,\alpha_2)\sim P(\alpha_2,\beta_2)\sim\\
&\quad\dots \sim (\beta_{k-1},\alpha_k)\sim P(\alpha_k,\beta_k)\\
q': P(\alpha_1',\beta_1')&\sim (\beta_1',\alpha_2')\sim P(\alpha_2',\beta_2')\sim\\
&\quad\dots \sim (\beta_{k-1}',\alpha_k')\sim P(\alpha_k',\beta_k')
\end{align*}
in the extended $\set{B}_{dag}$ where $\alpha_1=\alpha_1'$, $\beta_k=\beta_k'$, and $\alpha_i,\beta_i$ are in the same SCC $\set{G}_{b_i}$ while $\alpha_i',\beta_i'$ in $\set{G}_{b_i'}$. Thus it suffices to prove that $p$ is equivalent to $P(u_1,\alpha_1)\sim q\sim P(\beta_k,v_k)$ while $p'$ equivalent to $P(u_1',\alpha_1')\sim q'\sim P(\beta_k',v_k')$. (Recall that $u_1'=u_1$, etc.) Since the proofs for them are essentially identical, we only consider $p$.

In fact, $P(u_1,\alpha_1)\sim q\sim P(\beta_k,v_k)$ can be equivalently expressed as
\begin{align*}
    &P(u_1,v_1)\sim P(v_1,\beta_1)\sim(\beta_1,\alpha_2)\sim P(\alpha_2,u_2)\\
    &\quad \sim P(u_2,v_2)\sim P(v_2,\beta_2)\sim\dots \\
    &\qquad 
    \sim P(\beta_{k-1},\alpha_k)\sim(\alpha_k,u_k)\sim P(u_k,v_k).
\end{align*}
In other words, we split $P(\alpha_i,\beta_i)$ into $P(\alpha_i,u_i)\sim P(u_i,v_i)\sim P(v_i,\beta_i)$ for $i=2,\dots,k-1$. However, our lemma just states that
$$
P(v_i,\beta_i)\sim (\beta_i,\alpha_{i+1})
$$
is equivalent to
$$
(v_i,u_{i+1})\sim P(u_{i+1},\alpha_{i+1}).
$$
Thus by series of merge operations, $P(u_1,\alpha_1)\sim q\sim P(\beta_k,v_k)$ can be shown to be equivalent to
\begin{align*}
    &P(u_1,v_1)\sim (v_1,u_2)\sim P(u_2,\alpha_2)\sim P(\alpha_2,u_2)\\
    &\quad \sim P(u_2,v_2)\sim P(v_2,u_3)\sim\dots\\
    &\qquad
    \sim P(u_k,\alpha_k)\sim P(\alpha_k,u_k)\sim P(u_k,v_k),
\end{align*}
which is clearly $p$ by cancelling all consecutive $P$'s. \qed

\subsection{Proof of Proposition 3}

First note that in fact the bound $|\set{V}||\set{E}|$ theorem 3.1 can be improved to $(|\set{V}|-1)|\set{E}|$ since $|\set{P}|\leq |\set{V}|-1$ all time.

In this way the size of $\set{B}_i$ is bounded by $|\set{E}(\set{B}_i)|(|\set{V}(\set{B}_i)|-1)$. Suppose there are $k$ strongly connected components in $\set{G}$ and $c$ edges across different SCCs. Then there are at most $c$ edges in $\bigcup_{i,j}\set{B}_{ij}$ since the edge number of a spanning tree is less than that of vertices by 1. Notice $c$ is also the edge number of contracted graph $\set{G}_{dag}$. Hence for the $\set{G}_{dag}$, there are would be at most $(k-1)\times c$ items in $\set{B}_{dag}$. Also observe that each SCC can have at most $|\set{V}|-k+1$ vertices when there are $k$ SCCs. So the size of $\set{B}$ would be bounded by
\begin{align*}
    &(k-1)c+c+(|\set{V}|-k)|\set{E}|\\
    \leq & k|\set{E}|+(|\set{V}|-k)|\set{E}|\\
    = & |\set{V}||\set{E}|
\end{align*}
\qed

\begin{figure*}
\footnotesize{
\setlength\tabcolsep{0.8pt}
\begin{tabular}{cccccccc} \hline
Source & Target & Congealing & RASL & CollectionFlow & DSP & FlowWeb & Ours \\
\includegraphics[width=0.092\textwidth]{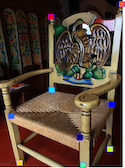} &
\includegraphics[width=0.092\textwidth]{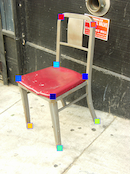} &
\includegraphics[width=0.092\textwidth]{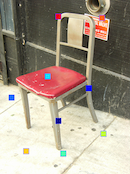} &
\includegraphics[width=0.092\textwidth]{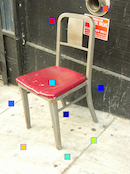} &
\includegraphics[width=0.092\textwidth]{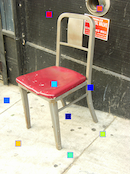} &
\includegraphics[width=0.092\textwidth]{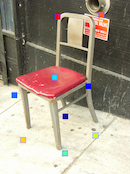} &
\includegraphics[width=0.092\textwidth]{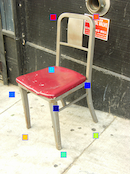} &
\includegraphics[width=0.092\textwidth]{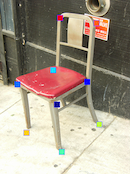}  \\
\includegraphics[width=0.122\textwidth]{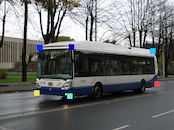} &
\includegraphics[width=0.122\textwidth]{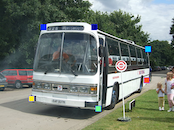} &
\includegraphics[width=0.122\textwidth]{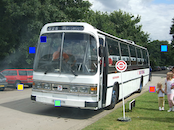} &
\includegraphics[width=0.122\textwidth]{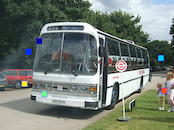} &
\includegraphics[width=0.122\textwidth]{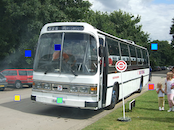} &
\includegraphics[width=0.122\textwidth]{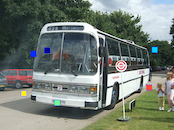} &
\includegraphics[width=0.122\textwidth]{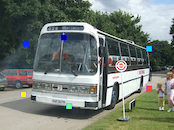} &
\includegraphics[width=0.122\textwidth]{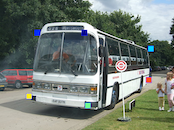} \\
\includegraphics[width=0.092\textwidth]{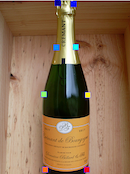} &
\includegraphics[width=0.092\textwidth]{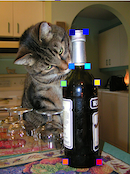} &
\includegraphics[width=0.092\textwidth]{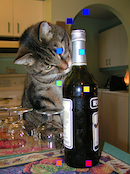} &
\includegraphics[width=0.092\textwidth]{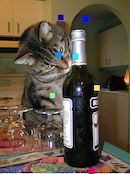} &
\includegraphics[width=0.092\textwidth]{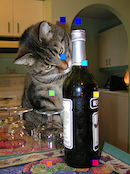} &
\includegraphics[width=0.092\textwidth]{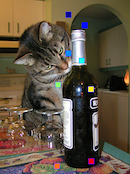} &
\includegraphics[width=0.092\textwidth]{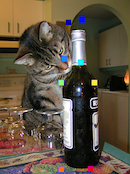} &
\includegraphics[width=0.092\textwidth]{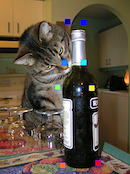} \\
\includegraphics[width=0.122\textwidth]{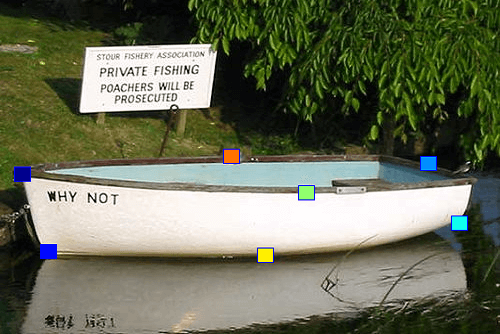} &
\includegraphics[width=0.122\textwidth]{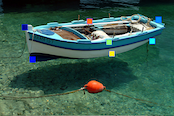} &
\includegraphics[width=0.122\textwidth]{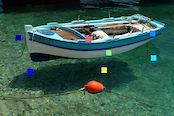} &
\includegraphics[width=0.122\textwidth]{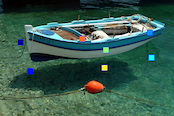} &
\includegraphics[width=0.122\textwidth]{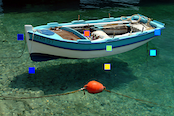} &
\includegraphics[width=0.122\textwidth]{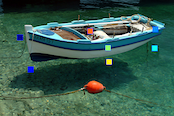} &
\includegraphics[width=0.122\textwidth]{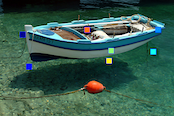} &
\includegraphics[width=0.122\textwidth]{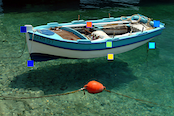} \\
\includegraphics[width=0.122\textwidth]{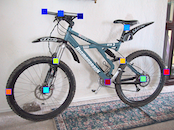} &
\includegraphics[width=0.122\textwidth]{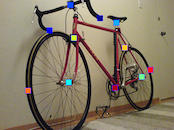} &
\includegraphics[width=0.122\textwidth]{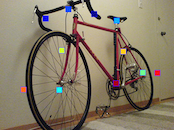} &
\includegraphics[width=0.122\textwidth]{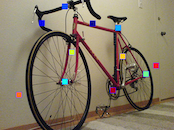} &
\includegraphics[width=0.122\textwidth]{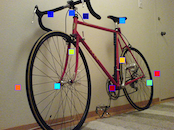} &
\includegraphics[width=0.122\textwidth]{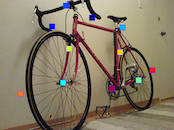} &
\includegraphics[width=0.122\textwidth]{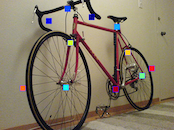} &
\includegraphics[width=0.122\textwidth]{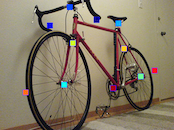} 
\end{tabular}}
\caption{\small{Visual comparison between our approach and state-of-the-art approaches. This figure is best viewed in color, zoomed in.}}
\vspace{-0.15in}
\label{Figure:Dense:Image:Flow:QualitativeII}
\end{figure*}
\section{Additional Details of Joint Dense Image Map}
\label{Section:Details:Joint:Dense:Image:Map}

\subsection{Training Details}

We applied ADAM~\cite{DBLP:journals/corr/KingmaB14} to solve the following optimization problem for predicting dense image correspondences. 
\begin{equation}
\min\limits_{\theta} \quad 
\sum\limits_{(i,j)\in \set{E}} \|f^{\theta}_{ij}-f_{ij}^{\init}\|_1 + \lambda \sum\limits_{(p,q)\in \set{B}} \|f_{p}^{\theta}-f_{q}^{\theta}\|_{\set{F}}^2
\label{Eq:FMap:Opt2}
\end{equation}
We initialize $f^{\theta}$ by directly fitting it to the input image flows between pairs of images. We then impose the path-invariance regularization term to improve the network flow. 

\subsection{More Qualitative Evaluations}

Figure~\ref{Figure:Dense:Image:Flow:QualitativeI} and Figure~\ref{Figure:Dense:Image:Flow:QualitativeII} provide more qualitative evaluations of our approach on the PASCAL Rigid categories. Besides the two categories shown in the main paper (Car and Aeroplane), we pick one example from the remaining 10 rigid categories. Note that our approach is consistently better than all baseline approaches. 

\subsection{Comparison to \cite{DBLP:conf/cvpr/ZhouKAHE16}}

We run additional experiments to compare our approach
against \cite{DBLP:conf/cvpr/ZhouKAHE16} (using DSP as input) on the Car dataset, where we used 1000 additional synthetic images. Table~\ref{tab:my_label} shows the performance of different baselines. Our approach is superior to \cite{DBLP:conf/cvpr/ZhouKAHE16}. The improvement comes from using the network to propagate learned flows between similar images.
Note that \cite{DBLP:conf/cvpr/ZhouKAHE16} essentially enforces consistency among selected 4-cycles (two synthetic and two real), so its performance is similar to Ours-Dense, which involves 3-cycles. 

\begin{table}[t]
\centering
\begin{tabular}{c|c|c|c|c}
         & Ours (Real-only) & \cite{DBLP:conf/cvpr/ZhouKAHE16} & Ours-Dense& Ours\\\hline
 PCK        & 0.65 &0.59 & 0.60&0.68
\end{tabular}
    \caption{Additional comparison to~\cite{DBLP:conf/cvpr/ZhouKAHE16}.}
    \label{tab:my_label}
\end{table}
\section{Additional Details of 3D Semantic Scene Segmentation}
\label{Section:Detail:Network:3D:Understanding}
\subsection{Network Architecture and Training Details}
For point cloud semantic segmentation network, we follow the same configuration from PointNet++~\cite{DBLP:journals/corr/QiYSG17}. For voxel semantic segmentation network, we use the same network architecture proposed in 3D U-Net\cite{cciccek20163d}. To generate training data from ScanNet scenes, following PointNet++~\cite{DBLP:journals/corr/QiYSG17}, we sample 1.5m by 1.5m by 3m cubes from the initial scenes. We sample such training cubes on the fly and randomly rotate each sample along the up-right axis. 
During test time, we split the test scene into smaller cubes first, and then merge label prediction in all the cubes from the same scene. Note that this is done for the prediction using each 3D representation in isolation. 

We applied ADAM~\cite{DBLP:journals/corr/KingmaB14} to solve the optimization problem for predicting semantic labels in 3D scenes. We first initialize network parameters using the pre-trained weight on labeled data, and then impose the path-invariance regularization term to improve the network performance. 
\subsection{More Quantitative Evaluations}
Table \ref{Figure:Map:Quantitative} shows per-class semantic voxel label prediction accuracy on ScanNet~\cite{dai2017scannet} test scenes. Compared to baseline methods, our approach shows consistently better performance compared to using 8\% labeled data, and competitive results compared to using 30\% and 100\% labeled data, especially on frequently appeared classes, such as floor, wall, chair, sofa, and etc.
\begin{table*}
\centering
\footnotesize{
\setlength\tabcolsep{1.07pt}
\begin{tabular}{c|cccccccccccccccccccc|c|}
 & Floor & Wall & Chair & Sofa & Table & Door & Cabinet & Bed & Desk & Toilet & Sink & Window & Picture & BookSh & Curtain & ShowerC & Counter & Fridge & Bathtub & OtherF & Total\\\hline
 Weight & 35.7 & 38.8 & 3.8 & 2.5 & 3.3 & 2.2 & 2.4 & 2.0 & 1.7 & 0.2 & 0.2 & 0.4 & 0.2 & 1.6 & 0.7 & 0.04 & 0.6 & 0.3 & 0.2 & 2.9 & -  \\\hline
 \multirow{4}{*}{PCI}
& 90.9 & 98.1 & 58.4 & 45.4 & 40.2 & 47.4 & 36.4 & 62.8 & 21.8 & 35.4 & 32.0 & 16.7 & 21.5 & 0.0 & 0.0 & 1.3 & 0.0 & 0.0 & 19.7 & 9.6 & 79.2\\
& 93.3 & 98.4 & 70.3 & 54.8 & 50.0 & 49.2 & 80.9 & 87.1 & 18.4 & 83.7 & 58.9 & 8.4 & 0.2 & 1.0 & 1.8 & 2.9 & 3.7 & 0.0 & 13.0 & 5.7 & 82.3\\
& 88.0 & 97.8 & 76.3 & 62.7 & 19.9 & 63.5 & 65.5 & 59.7 & 52.5 & 63.9 & 76.2 & 17.4 & 27.1 & 17.0 & 12.2 & 56.1 & 0.0 & 0.0 & 25.7 & 22.0 & 80.8\\
& 90.8 & 98.2 & 78.0 & 67.5 & 42.8 & 74.8 & 79.6 & 79.8 & 58.2 & 78.0 & 82.1 & 53.1 & 42.3 & 12.1 & 28.2 & 70.0 & 52.7 & 0.0 & 37.3 & 18.7 & 84.2\\\hline
  \multirow{4}{*}{PCII}
& 91.5 & 97.2 & 49.4 & 32.2 & 32.4 & 44.3 & 30.8 & 70.1 & 24.9 & 45.0 & 35.0 & 29.2 & 23.9 & 0.0 & 10.6 & 1.1 & 0.0 & 0.0 & 18.0 & 10.0 & 78.3\\
& 94.9 & 98.4 & 65.0 & 58.1 & 48.0 & 41.7 & 65.4 & 89.6 & 31.2 & 81.0 & 62.9 & 4.6 & 4.6 & 0.0 & 0.4 & 3.7 & 0.0 & 0.0 & 17.5 & 4.5 & 82.5 \\
& 90.8 & 98.5 & 74.4 & 54.6 & 34.4 & 49.3 & 46.7 & 77.3 & 39.3 & 74.8 & 71.9 & 22.8 & 35.6 & 0.0 & 0.0 & 24.8 & 0.0 & 0.0 & 25.4 & 11.7 & 81.9\\
& 92.8 & 98.0 & 86.4 & 64.2 & 29.8 & 55.0 & 59.2 & 75.3 & 37.6 & 86.5 & 67.6 & 9.3 & 25.3 & 23.5 & 19.0 & 46.6 & 43.1 & 0.0 & 25.0 & 13.7 & 83.3\\\hline
  \multirow{4}{*}{PCIII}
& 92.7 & 96.7 & 73.3 & 52.9 & 16.7 & 36.4 & 1.3 & 55.7 & 12.1 & 27.0 & 27.1 & 16.6 & 11.5 & 0.0 & 0.2 & 8.9 & 0.0 & 0.0 & 15.0 & 1.6 & 78.4\\
& 93.7 & 98.1 & 71.4 & 58.9 & 50.0 & 54.4 & 59.9 & 74.8 & 30.6 & 82.8 & 65.1 & 10.6 & 1.6 & 1.4 & 0.8 & 21.5 & 0.0 & 0.0 & 20.3 & 8.7 & 82.3 \\
& 90.8 & 98.5 & 74.4 & 54.6 & 34.4 & 49.3 & 46.7 & 77.3 & 39.3 & 74.8 & 71.9 & 22.8 & 35.6 & 0.0 & 0.0 & 24.8 & 0.0 & 0.0 & 25.4 & 11.7 & 81.2\\
& 90.4 & 97.6 & 76.1 & 65.0 & 45.5 & 80.6 & 70.9 & 75.3 & 32.4 & 82.0 & 73.9 & 48.0 & 49.8 & 13.5 & 16.9 & 64.4 & 46.7 & 0.0 & 42.0 & 13.0 & 83.4\\\hline
  \multirow{4}{*}{VOLI}
& 93.4 & 97.3 & 71.9 & 68.0 & 16.2 & 0.2 & 0.0 & 58.1 & 34.3 & 25.1 & 3.2 & 0.0 & 0.0 & 0.0 & 0.0 & 0.0 & 0.0 & 0.0 & 18.3 & 8.6 & 78.7\\
& 93.5 & 97.6 & 70.7 & 61.2 & 55.7 & 39.1 & 55.0 & 76.7 & 11.5 & 81.3 & 68.8 & 0.3 & 2.3 & 2.2 & 0.0 & 2.0 & 0.0 & 0.0 & 16.8 & 10.2 & 81.6\\
& 94.0 & 97.6 & 68.0 & 68.2 & 16.7 & 41.2 & 0.0 & 75.1 & 0.0 & 70.2 & 30.4 & 0.0 & 0.0 & 0.4 & 0.0 & 0.0 & 0.0 & 0.0 & 24.9 & 6.9 & 80.3\\
& 92.5 & 97.5 & 74.2 & 67.2 & 25.0 & 55.0 & 59.5 & 62.9 & 0.0 & 85.4 & 0.0 & 3.9 & 38.5 & 0.4 & 0.0 & 0.0 & 42.5 & 0.0 & 37.8 & 14.2 & 81.9\\\hline
   \multirow{4}{*}{VOLII}
& 94.8 & 97.5 & 56.0 & 0.0 & 42.3 & 19.8 & 28.3 & 57.3 & 9.7 & 0.0 & 0.0 & 0.0 & 0.0 & 0.0 & 0.0 & 0.0 & 0.0 & 0.0 & 13.5 & 5.1 & 77.4\\
& 92.8 & 97.7 & 69.6 & 0.0 & 53.8 & 31.6 & 66.4 & 68.2 & 11.4 & 77.3 & 0.0 & 0.0 & 0.0 & 0.0 & 1.1 & 0.0 & 0.0 & 0.0 & 19.8 & 9.8 & 79.0\\
& 92.5 & 98.1 & 62.4 & 54.4 & 15.3 & 50.0 & 0.0 & 59.1 & 0.0 & 74.5 & 61.9 & 0.0 & 0.0 & 0.0 & 0.0 & 0.0 & 0.0 & 0.0 & 39.0 & 9.3 & 79.5\\
& 91.0 & 96.9 & 68.4 & 60.5 & 31.4 & 59.1 & 70.0 & 81.2 & 0.0 & 86.3 & 0.0 & 11.1 & 0.0 & 0.0 & 1.4 & 0.0 & 0.0 & 0.0 & 50.1 & 15.1 & 81.5\\\hline
\end{tabular}}
\caption{\small{Per-class semantic voxel label prediction accuracy on ScanNet test scenes. All numbers are in percentages. The first row indicates the percentages of each class in all test scenes, and then for 4 rows in each representation, we show the per-class prediction accuracy in 4 configurations: 8\% Label, 8\% Label + 92\%Unlabel, 30\% Label and 100\% Label. (BookSh, ShowerC and OtherF are short Bookshelf, Shower Curtain and Other furniture, respectively.)}}
\label{Figure:Map:Quantitative}
\vspace{-0.15in}
\end{table*}
\subsection{More Qualitative Evaluations}
\begin{figure*}[htb]
\centering
\setlength\tabcolsep{2pt}
\begin{tabular}{l | l l l l}
Ground Truth & 8\% Label & 30\% Label & 100\% Label & 8\% Label + 92\%Unlabel \\\hline
\multirow{2}{*}{
\includegraphics[width=0.4\columnwidth]{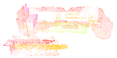}}
&
\multicolumn{1}{l}{\includegraphics[width=0.4\columnwidth]{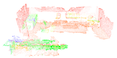}}
&
\multicolumn{1}{l}{\includegraphics[width=0.4\columnwidth]{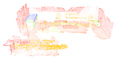}}
&
\multicolumn{1}{l}{\includegraphics[width=0.4\columnwidth]{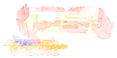}}
&
\multicolumn{1}{l}{\includegraphics[width=0.4\columnwidth]{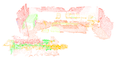}}\\
&
\multicolumn{1}{l}{\includegraphics[width=0.4\columnwidth]{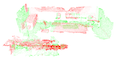}}
&
\multicolumn{1}{l}{\includegraphics[width=0.4\columnwidth]{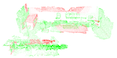}}
&
\multicolumn{1}{l}{\includegraphics[width=0.4\columnwidth]{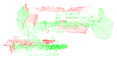}}
&
\multicolumn{1}{l}{\includegraphics[width=0.4\columnwidth]{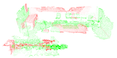}}\\\hline
\multirow{2}{*}{
\includegraphics[width=0.4\columnwidth]{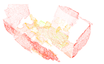}}
&
\multicolumn{1}{l}{\includegraphics[width=0.4\columnwidth]{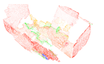}}
&
\multicolumn{1}{l}{\includegraphics[width=0.4\columnwidth]{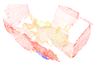}}
&
\multicolumn{1}{l}{\includegraphics[width=0.4\columnwidth]{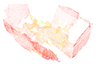}}
&
\multicolumn{1}{l}{\includegraphics[width=0.4\columnwidth]{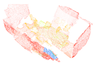}}\\
&
\multicolumn{1}{l}{\includegraphics[width=0.4\columnwidth]{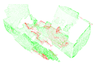}}
&
\multicolumn{1}{l}{\includegraphics[width=0.4\columnwidth]{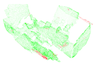}}
&
\multicolumn{1}{l}{\includegraphics[width=0.4\columnwidth]{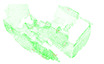}}
&
\multicolumn{1}{l}{\includegraphics[width=0.4\columnwidth]{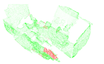}}\\\hline
\multirow{2}{*}{
\includegraphics[width=0.4\columnwidth]{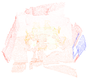}}
&
\multicolumn{1}{l}{\includegraphics[width=0.4\columnwidth]{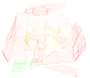}}
&
\multicolumn{1}{l}{\includegraphics[width=0.4\columnwidth]{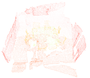}}
&
\multicolumn{1}{l}{\includegraphics[width=0.4\columnwidth]{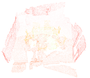}}
&
\multicolumn{1}{l}{\includegraphics[width=0.4\columnwidth]{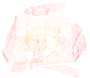}}\\
&
\multicolumn{1}{l}{\includegraphics[width=0.4\columnwidth]{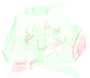}}
&
\multicolumn{1}{l}{\includegraphics[width=0.4\columnwidth]{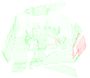}}
&
\multicolumn{1}{l}{\includegraphics[width=0.4\columnwidth]{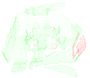}}
&
\multicolumn{1}{l}{\includegraphics[width=0.4\columnwidth]{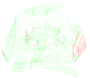}}\\\hline
\multirow{2}{*}{
\includegraphics[width=0.4\columnwidth]{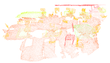}}
&
\multicolumn{1}{l}{\includegraphics[width=0.4\columnwidth]{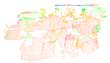}}
&
\multicolumn{1}{l}{\includegraphics[width=0.4\columnwidth]{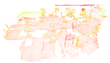}}
&
\multicolumn{1}{l}{\includegraphics[width=0.4\columnwidth]{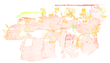}}
&
\multicolumn{1}{l}{\includegraphics[width=0.4\columnwidth]{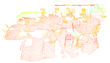}}\\
&
\multicolumn{1}{l}{\includegraphics[width=0.4\columnwidth]{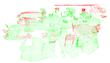}}
&
\multicolumn{1}{l}{\includegraphics[width=0.4\columnwidth]{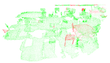}}
&
\multicolumn{1}{l}{\includegraphics[width=0.4\columnwidth]{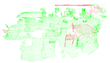}}
&
\multicolumn{1}{l}{\includegraphics[width=0.4\columnwidth]{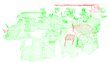}}\\\hline
\multirow{2}{*}{
\includegraphics[width=0.4\columnwidth]{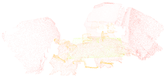}}
&
\multicolumn{1}{l}{\includegraphics[width=0.4\columnwidth]{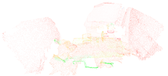}}
&
\multicolumn{1}{l}{\includegraphics[width=0.4\columnwidth]{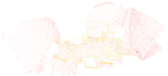}}
&
\multicolumn{1}{l}{\includegraphics[width=0.4\columnwidth]{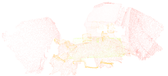}}
&
\multicolumn{1}{l}{\includegraphics[width=0.4\columnwidth]{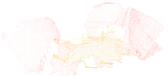}}\\
&
\multicolumn{1}{l}{\includegraphics[width=0.4\columnwidth]{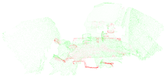}}
&
\multicolumn{1}{l}{\includegraphics[width=0.4\columnwidth]{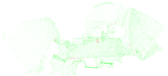}}
&
\multicolumn{1}{l}{\includegraphics[width=0.4\columnwidth]{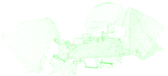}}
&
\multicolumn{1}{l}{\includegraphics[width=0.4\columnwidth]{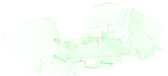}}\\\hline
\end{tabular}
\caption{\small{Qualitative comparisons of 3D semantic segmentation results on ScanNet~\protect\cite{dai2017scannet}.} Each row represents one testing instance, where ground truth and top sub-row show prediction for 21 classes and bottom sub-row only shows correctly labeled points. (Green indicates correct predictions, while red indicates false predictions.) This figure is best viewed in color, zoomed in. }
\label{Figure:Map:Network:3D:Understanding_more}
\vspace{-0.15in}
\end{figure*}

Figure~\ref{Figure:Map:Network:3D:Understanding_more} presents more qualitative comparisons between our approach and baselines. Consistently, using 8\% labeled data and 92\% unlabeled data, our approach achieved competing performance as using 30\% to 100\% labeled data when trained on each individual representation, and better performance as using 8\% labeled data.
\begin{figure*}[htb]
\centering
\setlength\tabcolsep{2pt}
\begin{tabular}{l | l l l l}
Ground Truth & 8\% Label & 30\% Label & 100\% Label & 8\% Label + 92\%Unlabel \\\hline
\multirow{2}{*}{
\includegraphics[width=0.4\columnwidth]{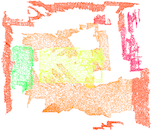}}
&
\multicolumn{1}{l}{\includegraphics[width=0.4\columnwidth]{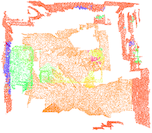}}
&
\multicolumn{1}{l}{\includegraphics[width=0.4\columnwidth]{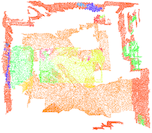}}
&
\multicolumn{1}{l}{\includegraphics[width=0.4\columnwidth]{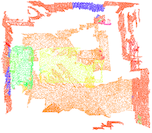}}
&
\multicolumn{1}{l}{\includegraphics[width=0.4\columnwidth]{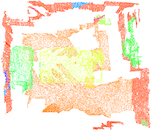}}\\
&
\multicolumn{1}{l}{\includegraphics[width=0.4\columnwidth]{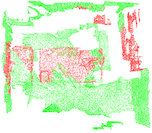}}
&
\multicolumn{1}{l}{\includegraphics[width=0.4\columnwidth]{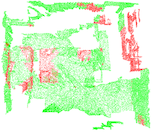}}
&
\multicolumn{1}{l}{\includegraphics[width=0.4\columnwidth]{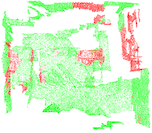}}
&
\multicolumn{1}{l}{\includegraphics[width=0.4\columnwidth]{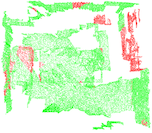}}\\\hline
\multirow{2}{*}{
\includegraphics[width=0.4\columnwidth]{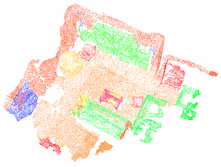}}
&
\multicolumn{1}{l}{\includegraphics[width=0.4\columnwidth]{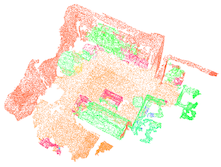}}
&
\multicolumn{1}{l}{\includegraphics[width=0.4\columnwidth]{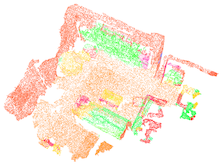}}
&
\multicolumn{1}{l}{\includegraphics[width=0.4\columnwidth]{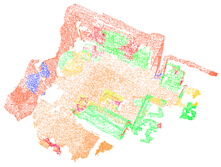}}
&
\multicolumn{1}{l}{\includegraphics[width=0.4\columnwidth]{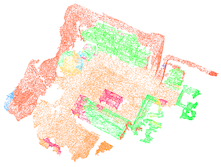}}\\\cline{2-5}
&
\multicolumn{1}{l}{\includegraphics[width=0.4\columnwidth]{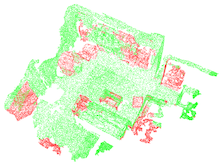}}
&
\multicolumn{1}{l}{\includegraphics[width=0.4\columnwidth]{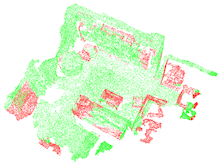}}
&
\multicolumn{1}{l}{\includegraphics[width=0.4\columnwidth]{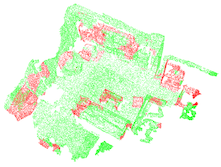}}
&
\multicolumn{1}{l}{\includegraphics[width=0.4\columnwidth]{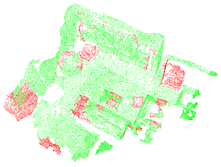}}\\\hline
\multirow{2}{*}{
\includegraphics[width=0.4\columnwidth]{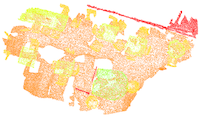}}
&
\multicolumn{1}{l}{\includegraphics[width=0.4\columnwidth]{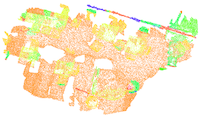}}
&
\multicolumn{1}{l}{\includegraphics[width=0.4\columnwidth]{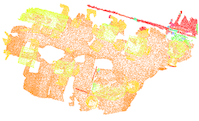}}
&
\multicolumn{1}{l}{\includegraphics[width=0.4\columnwidth]{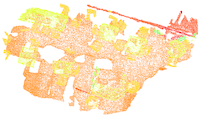}}
&
\multicolumn{1}{l}{\includegraphics[width=0.4\columnwidth]{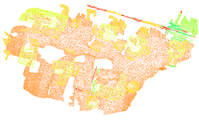}}\\\cline{2-5}
&
\multicolumn{1}{l}{\includegraphics[width=0.4\columnwidth]{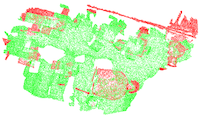}}
&
\multicolumn{1}{l}{\includegraphics[width=0.4\columnwidth]{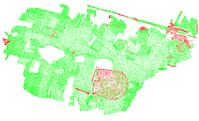}}
&
\multicolumn{1}{l}{\includegraphics[width=0.4\columnwidth]{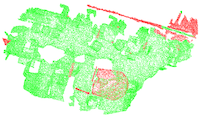}}
&
\multicolumn{1}{l}{\includegraphics[width=0.4\columnwidth]{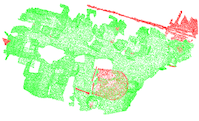}}\\\hline
\multirow{2}{*}{
\includegraphics[width=0.4\columnwidth]{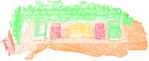}}
&
\multicolumn{1}{l}{\includegraphics[width=0.4\columnwidth]{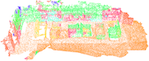}}
&
\multicolumn{1}{l}{\includegraphics[width=0.4\columnwidth]{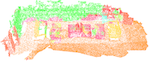}}
&
\multicolumn{1}{l}{\includegraphics[width=0.4\columnwidth]{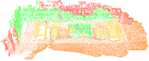}}
&
\multicolumn{1}{l}{\includegraphics[width=0.4\columnwidth]{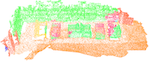}}\\
&
\multicolumn{1}{l}{\includegraphics[width=0.4\columnwidth]{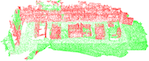}}
&
\multicolumn{1}{l}{\includegraphics[width=0.4\columnwidth]{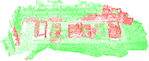}}
&
\multicolumn{1}{l}{\includegraphics[width=0.4\columnwidth]{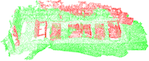}}
&
\multicolumn{1}{l}{\includegraphics[width=0.4\columnwidth]{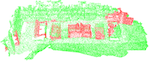}}\\\hline
\multirow{2}{*}{
\includegraphics[width=0.4\columnwidth]{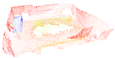}}
&
\multicolumn{1}{l}{\includegraphics[width=0.4\columnwidth]{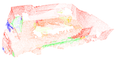}}
&
\multicolumn{1}{l}{\includegraphics[width=0.4\columnwidth]{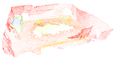}}
&
\multicolumn{1}{l}{\includegraphics[width=0.4\columnwidth]{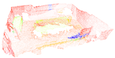}}
&
\multicolumn{1}{l}{\includegraphics[width=0.4\columnwidth]{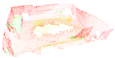}}\\
&
\multicolumn{1}{l}{\includegraphics[width=0.4\columnwidth]{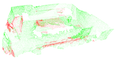}}
&
\multicolumn{1}{l}{\includegraphics[width=0.4\columnwidth]{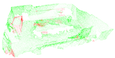}}
&
\multicolumn{1}{l}{\includegraphics[width=0.4\columnwidth]{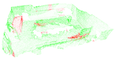}}
&
\multicolumn{1}{l}{\includegraphics[width=0.4\columnwidth]{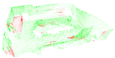}}\\\hline
\end{tabular}
\vspace{-0.15in}
\end{figure*}

\section{Additional Details of Joint Shape Matching}
\label{Section:Details:Joint:Shape:Matching}

\subsection{Training Details}

We applied ADAM~\cite{DBLP:journals/corr/KingmaB14} to solve the following optimization problem:
\begin{equation}
\sum\limits_{(i,j)\in \set{E}}\|X_{ij}-X_{ij}^{\init}\|_1 + \lambda \sum\limits_{(p,q)\in \set{B}}\|X_{p} - X_{q}\|_{\set{F}}^2    
\end{equation}
Initially, we set $X_{ij} = X_{ij}^{\init}$. We also tried reweighted non-linear least squares and used Gauss-Newton optimization to solve the induced non-linear least square problem (we used conjugate gradient to solve the induced linear system) . We found that the optimal solutions of both approaches are similar, suggesting both of them reached a strong local minimum. Computationally, we find the ADAM optimizer to be more efficient. 

\subsection{Annotated Feature Points}

\subsubsection{SHREC07}

We used annotated feature points provided by~\cite{Kim:2011:BIM}. The number of key points per category range from 11 (e.g. Plane) to 36 (Human).

\subsubsection{ShapeCoSeg}

Note that the models in ShapeCoSeg~\cite{Wang:2012:ACS} are originally associated with annotations of semantic segments. Such annotations, however, are not ideal for establishing dense correspondences. To address this issue, we employed AMT to annotate semantic feature correspondences across the entire dataset. Note that in some cases, the feature correspondences are not purely based on 1-1 correspondences (e.g., multiple handles). When performing experimental evaluation, we evaluate the geodesic error to the closest feature point of the same type for experimental evaluation.

\end{document}